\documentclass{article}

 \usepackage[preprint]{neurips_2026}

\usepackage[utf8]{inputenc} 
\usepackage[T1]{fontenc}    
\usepackage{hyperref}       
\usepackage{url}            
\usepackage{booktabs}       
\usepackage{amsfonts}       
\usepackage{nicefrac}       
\usepackage{microtype}      
\usepackage{xcolor}         

\usepackage[pdftex]{graphicx}
\PassOptionsToPackage{options}{natbib}
\usepackage{subcaption}
\usepackage{amsmath,amssymb,amsfonts}

\usepackage[utf8]{inputenc} 
\usepackage[T1]{fontenc}    
\usepackage{algorithm}
\usepackage{algpseudocode}
\usepackage{multirow}
\usepackage{wrapfig}   
\usepackage{caption}   
\usepackage{makecell}
\usepackage{xcolor}
\definecolor{lightblue}{RGB}{123,173,255} 
\newcommand{\revB}[1]{\textcolor{lightblue}{#1}}   
\usepackage[table]{xcolor} 
\usepackage{colortbl}      
% ...
\title{EVA: Bridging Performance and Human Alignment in Hard-Attention Vision Models for Image Classification}

\author{%
	Pengcheng Pan \quad Yonekura Shogo \quad Kuniyoshi Yasuo\\
	The University of Tokyo\\
	\texttt{\{pan,yonekura,kuniyoshi\}@isi.imi.i.u-tokyo.ac.jp} 
}

\begin{document}

\maketitle

\begin{abstract}
  Optimizing vision models purely for classification accuracy can impose an alignment tax, degrading human-like scanpaths and limiting interpretability. We introduce EVA, a neuroscience-inspired hard-attention mechanistic testbed that makes the performance–human-likeness trade-off explicit and adjustable. EVA samples a small number of sequential glimpses using a minimal fovea–periphery representation with CNN-based feature extractor and integrates variance control and adaptive gating to stabilize and regulate attention dynamics. EVA is trained with the standard classification objective without gaze supervision. On CIFAR-10 with dense human gaze annotations, EVA improves scanpath alignment under established metrics such as DTW, NSS, while maintaining competitive accuracy. Ablations show that CNN-based feature extraction drives accuracy but suppresses human-likeness, whereas variance control and gating restore human-aligned trajectories with minimal performance loss. We further validate EVA’s scalability on ImageNet-100 and evaluate scanpath alignment on COCO-Search18 without COCO-Search18 gaze supervision or finetuning, where EVA yields human-like scanpaths on natural scenes without additional training. Overall, EVA provides a principled framework for trustworthy, human-interpretable active vision.
\end{abstract}

\section{Introduction}
\label{sec:intro}

Deploying vision systems in real-world settings requires not only strong task performance but also \emph{interpretability} and \emph{reliability}. When humans cannot anticipate why a model reaches a decision, trust and safe integration into high-stakes workflows remain limited. Popular post-hoc explanation tools such as saliency maps and Grad-CAM for CNNs and Vision Transformers can be brittle and occasionally misleading, raising concerns about whether they reflect the model's actual reasoning process \cite{CNN,ViT,Interpretable1,GradCAM}. This motivates a complementary direction: designing models whose internal decision process is inspectable by construction, rather than explained after the fact.

Human vision provides a natural blueprint for process-level interpretability. Due to biological constraints, perception is selective. Humans cannot process the entire visual field at high resolution simultaneously. Instead, they actively sample evidence through sequences of saccades and fixations, repositioning the fovea to gather informative glimpses \cite{attention1,Yarbus,Eye1}. Hard-attention models follow this principle by selecting discrete glimpses over time and learning fixation policies, often using reinforcement learning \cite{RAM,REINFORCE}. Their trajectories, or scanpaths, form an auditable record of what evidence was requested to support a decision. This makes scanpath behavior a natural axis for interpretability.

Modern vision research, however, predominantly optimizes for classification accuracy under full-image, parallel computation. In practice, this can impose an \emph{alignment tax}. As accuracy improves, scanpath behavior often becomes less human-like, weakening the interpretability promise of active sampling. Rather than treating this trade-off as inevitable, we argue that the performance and human-likeness tension should be treated as a first-class design variable and studied mechanistically. We evaluate scanpath alignment by comparing model fixations to human gaze statistics under established metrics such as NSS and DTW. We then ask which architectural mechanisms can reduce the alignment tax under the standard classification objective.

\begin{figure}[htbp]
	\centerline{\includegraphics[scale=0.47]{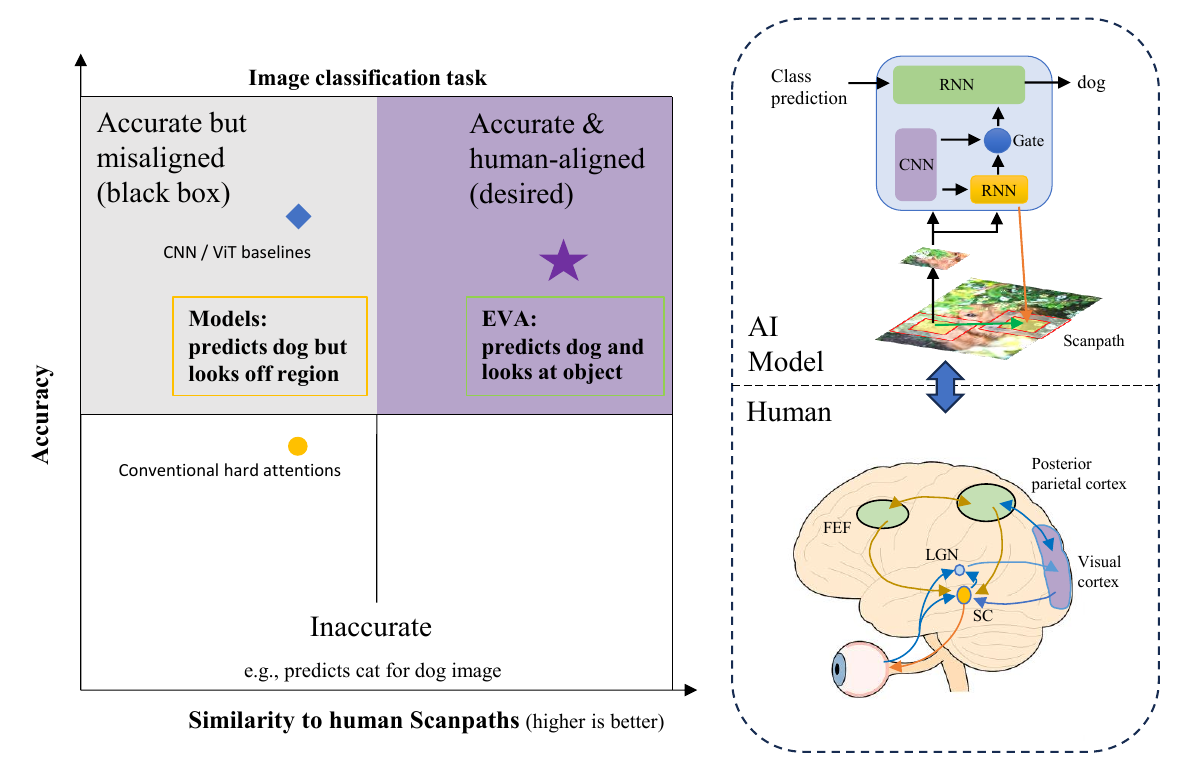}}
	\caption{EVA reduces the alignment tax through mechanism-aware design. \textbf{Left}: a schematic trade-off between accuracy and scanpath similarity to humans. EVA improves the trade-off relative to full-image baselines and prior hard-attention models. \textbf{Right}: EVA uses minimal fovea--periphery sensing with variance control and adaptive gating to regulate active sampling dynamics.}
	\label{fig01}
\end{figure}

We introduce \textbf{EVA}, a lightweight neuroscience-inspired hard-attention mechanistic testbed trained solely with classification labels. EVA is designed to make the alignment tax measurable and adjustable through architectural constraints, while keeping the training objective unchanged. EVA operates with a small number of sequential glimpse requests and combines a minimal fovea--periphery glimpse representation with a CNN-based feature extractor, a neuromodulator-inspired variance control mechanism that stabilizes stochastic attention under uncertainty, and an adaptive gating pathway that regulates information flow. Our goal is not biological fidelity. Instead, these components impose functional priors on how the model acquires evidence over time. We adopt a minimal two-level fovea--periphery design to maintain efficiency and to avoid confounds introduced by complex multi-scale designs \cite{DebiasCentralFixation}.

Our evaluation is structured to support mechanistic analysis, scalability testing, and cross-task transfer. On \textbf{CIFAR-10}, where dense human gaze annotations enable rigorous analysis, we conduct mechanistic ablations to decouple the sources of performance and alignment. We find that stronger CNN-based feature extraction primarily drives accuracy but tends to suppress scanpath human-likeness. In contrast, variance control and gating substantially restore human-aligned fixation trajectories with minimal performance loss. We then validate scalability on \textbf{ImageNet-100}. Finally, we evaluate scanpath alignment on \textbf{COCO-Search18} without using its gaze annotations or search supervision. We train EVA on COCO image classification using labels only, optimize the attention policy solely via classification reward, and compare the resulting scanpaths to human search-conditioned scanpaths on COCO-Search18 without finetuning. This setting tests whether an attention policy learned from classification reward can transfer as a human-aligned evidence acquisition strategy in a different behavioral regime.

\paragraph{Contributions.}
First, we formalize and empirically characterize an alignment tax in hard-attention vision. Accuracy-oriented improvements can systematically degrade scanpath human-likeness. Second, we introduce EVA as a mechanistic testbed that treats the performance and human-likeness trade-off as a first-class variable. EVA uses minimal fovea--periphery sensing together with variance control and adaptive gating as interpretable architectural mechanisms. Third, through ablations on CIFAR-10 and evaluations on ImageNet-100 and COCO-Search18, we show that EVA improves scanpath alignment without changing the standard training objective or using gaze supervision. EVA also scales to larger data and supports gaze-supervision-free cross-task scanpath evaluation. By aligning active sampling behavior with human gaze statistics, EVA enables a practical form of interpretability. Users can inspect where the model requested evidence and relate decisions to a human-understandable sequence of glimpses, supporting trustworthy and potentially human-in-the-loop active vision systems.

\section{Related Work}
\label{sec:related}

\subsection{Neuroscience-inspired Vision and Hard Attention}
Deep CNNs trained for object recognition have been used as functional models of the primate ventral stream, with empirical correspondences to neural responses and benchmarks that quantify neural alignment \cite{CNNVC,Xu,CNNVC2}. These models capture hierarchical feature processing but do not directly model active evidence acquisition through eye movements. Biological vision relies on recurrent loops and routing mechanisms that modulate information flow during perception, including attentional gating in thalamocortical circuits \cite{pulvinar,pulvinar2,biased1,attention3}. Hard-attention models move toward this active-sampling view by making decisions from a sequence of glimpses selected by an attention policy. RAM learns where to look using reinforcement learning signals such as REINFORCE, which introduces high-variance optimization challenges \cite{RAM,REINFORCE}. Subsequent work improved representation learning and training strategy to narrow the accuracy gap on more complex benchmarks, including designs that use multi-scale glimpses or stronger feature extractors and systems that scale to ImageNet with more effective training \cite{DRAM,saccader}. Related active-vision efforts also study uncertainty-driven exploration and multi-level controllers that separate perception and action \cite{PCRAM,MRAM}. EVA follows this line of work by introducing neuroscience-inspired components into a hard-attention system, not to claim biological fidelity, but to impose functional priors on how the model acquires evidence over time.

\subsection{Human Gaze and Scanpath Alignment}
Human gaze datasets enable evaluation of where and how models allocate attention. Saliency models and deep saliency networks predict fixation distributions for free viewing, but they often output static maps rather than sequential scanpaths \cite{saliency,attention5}. With datasets such as COCO-Search18 and Gaze-CIFAR10, recent methods explicitly predict scanpaths using powerful sequence models trained on eye-movement data \cite{COCO,COCO2,SemSS,gazecifar,DeepGaze,yang2024unify,Mondal_2023_CVPR,cartella2025modeling,xianyu:2024:gazexplain,review1}. These approaches treat gaze as a supervised target. In contrast, our work studies whether human-like scanpaths can emerge without gaze supervision in a task-driven hard-attention system. We treat the performance and human-likeness tension as a first-class variable, an alignment tax, and use EVA as a mechanistic testbed to identify architectural mechanisms that reduce this tax under the standard classification objective. We evaluate scanpath similarity under established metrics and test cross-task transfer by comparing EVA scanpaths to human search-conditioned scanpaths on COCO-Search18, while training the policy only via classification reward.

\section{Method}
\label{sec:method} 

\begin{figure}[htbp]
	\centerline{\includegraphics[scale=0.42]{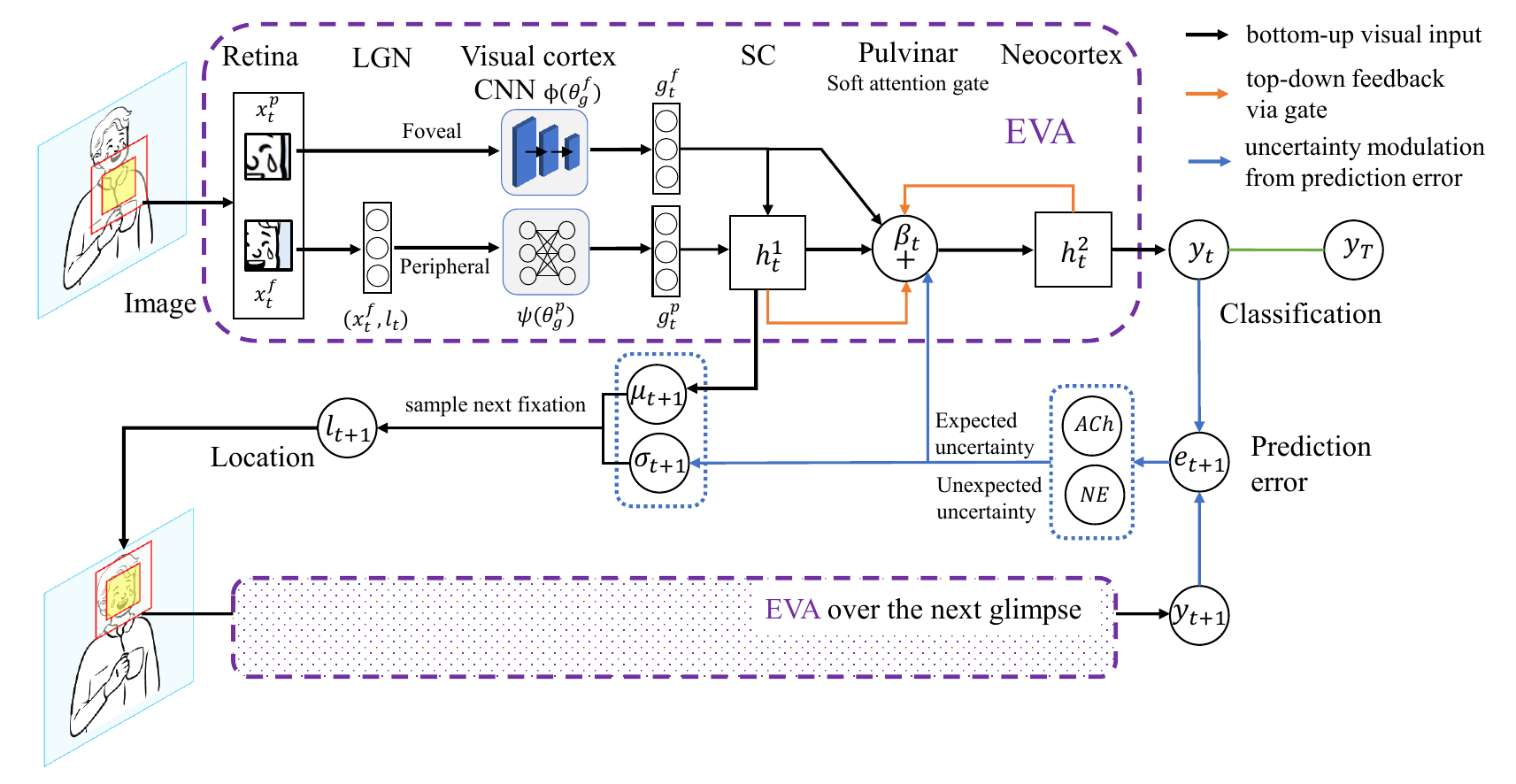}}
	\caption{EVA model architecture. \textbf{Top}: one glimpse step at time $t$. EVA combines a minimal fovea--periphery retina with a CNN feature extractor and a two-level recurrent backbone. The lower recurrent state updates the glimpse representation and controls the next fixation. An adaptive gate regulates information flow to the upper recurrent classifier. \textbf{Bottom}: a prediction-error signal modulates the fixation variance and the gate, regulating evidence acquisition dynamics. The components are motivated by functional motifs rather than biological fidelity.}
	\label{fig1}
\end{figure}

Hard-attention models emulate active vision by making decisions from a sequence of glimpses rather than processing the full image in parallel. A key practical challenge arises when pairing hard attention with strong feature extractors. Improving representation strength can raise classification accuracy while weakening the need for informative sequential sampling, leading to policy collapse and reduced scanpath human-likeness \cite{DebiasCentralFixation}. This behavior is consistent with the alignment tax studied in this paper.

We propose \textbf{EVA}, a neuroscience-inspired hard-attention architecture that makes this tension measurable and adjustable through mechanism-aware design. EVA can be trained end-to-end with classification labels only. It introduces three components that impose functional priors on how evidence is acquired over time. First, a CNN processes the foveal input to support accurate recognition. Second, a variance control mechanism modulates the stochasticity of fixation sampling based on prediction error to stabilize exploration. Third, an adaptive gating pathway regulates the influence of low-level sampling dynamics on high-level evidence integration. Together, these mechanisms help preserve meaningful sequential evidence acquisition while benefiting from stronger representations.

\subsection{Overview of RAM and MRAM}
RAM and its extension MRAM consist of four main modules: a \emph{glimpse module}, a \emph{core recurrent module}, a \emph{location module}, and an \emph{action module}. The glimpse module extracts features from a local image patch, mimicking foveal vision. The recurrent module integrates these glimpses over time. The location module predicts the next fixation, while the action module makes the classification decision.  
EVA adopts the two-level recurrent MRAM as backbone that separates sampling control and evidence integration. The lower RNN predicts fixation locations, while the upper RNN integrates sequential evidence for classification. At each timestep $t$, given a foveal crop $x^f_t$ and peripheral context $x^p_t$, the model computes:

\begin{alignat}{3}
	g_t &= f_g\!\left([x^f_t \| x^p_t], l_t; \theta_g\right), \quad &
	h_t^1 &= f_h^1(h_{t-1}^1, g_t; \theta_h^1), \quad &
	h_t^2 &= f_h^2(h_{t-1}^2, h_t^1; \theta_h^2), \label{RAM}\\
	l_{t+1} &\sim p(\cdot \mid f_l(h_t^1; \theta_l)), \quad &
	a_t &= f_a(h_t^2; \theta_a). &
\end{alignat}

Here $g_t$ is the glimpse representation, $h^1_t$ and $h^2_t$ are the lower and upper RNN states, $l_{t+1}$ is the fixation location sampled from a stochastic distribution, and $a_t$ is the classification output.

\subsection{CNN Feature Extractor for Fovea--Periphery Input}
EVA uses a CNN feature extractor $\phi(\cdot)$ for the foveal crop $x^f_t$ and a lightweight encoder $\psi(\cdot)$ for the peripheral view $x^p_t$ together with the current fixation $l_t$. We concatenate the two feature vectors to form the glimpse representation:
\[
g^f_t = \phi\!\left(x^f_t; \theta^f_g\right), \quad 
g^p_t = \psi\!\left(x^p_t, l_t; \theta^p_g\right), \quad 
\mathbf{s}_t = \bigl[g^f_t \|\; g^p_t  \bigr]. \tag{3}\label{cnn}
\]

The lower recurrent state integrates $\mathbf{s}_t$ to update the internal summary used for sampling control:
\[
h^{1}_{t} = f_{1}(h^{1}_{t-1}, \mathbf{s}_{t}). \tag{4}
\]

\subsection{Variance Control via Prediction Error}
The location module samples the next fixation from a Gaussian policy:
\[
l_{t+1} \sim \mathcal{N}(\mu(h_t^1), \sigma_t^2). \tag{5}
\]
RAM uses a fixed variance. EVA adapts $\sigma_t$ using a prediction-error driven signal that regulates exploration. This design is motivated by neuromodulatory control of uncertainty and exploration in biological systems \cite{ACh}, but EVA does not aim for biological fidelity.
We compute long-term and short-term exponential moving averages of the prediction error:
\[
\bar{e}^{(k)}_{t} = \tau_{k}\bar{e}^{(k)}_{t-1} + (1-\tau_{k}) e_t, \quad k\in\{\text{long},\text{short}\}, \tag{6}\label{tau}
\]
where $e_t$ is instantaneous error, and $\tau_{\text{long}}>\tau_{\text{short}}$. The uncertainty signal is
\[
u_t = \bigl| \bar{e}^{\text{s}}_{t} - \bar{e}^{\text{l}}_{t} \bigr|. \tag{7}
\]
Finally, $\sigma_t$ is bounded using:
\[
\sigma_t = \sigma_{\min} + (\sigma_{\max}-\sigma_{\min})\tanh(\alpha u_t). \tag{8}\label{sigma}
\]
Thus, high uncertainty increases fixation variance, encouraging exploration, while low uncertainty reduces variance and stabilizes sampling.

\subsection{Adaptive Gating}
EVA uses an adaptive gate that regulates information flow from the lower recurrent state $h^1_t$ to the upper recurrent state $h^2_t$. The gate depends on bottom-up and top-down signals and on the current sampling variance $\sigma_t$:
\begin{align}
	\beta_t &= \operatorname{clamp}\!\Bigl(
	(f_{td}(h^2_{t-1}) - \sigma_t) \odot (f_{bu}(h^1_{t-1}) + \sigma_t),\;0,1\Bigr), \tag{9}\\
	\bar{\beta}_t &= \gamma\,\bar{\beta}_{t-1} + (1-\gamma)\beta_t. \tag{10}\label{beta}
\end{align}
Both $f_{td}$ and $f_{bu}$ are implemented as single-layer MLPs with ReLU activations. The smoothed gate $\bar{\beta}_t$ provides a stable control signal over time.
We implement gated integration using an attention-style update \cite{NIPS2017Attention}. The upper state forms a query, and the lower features form key and value vectors:
\begin{alignat}{2}
	Q_t &= W_Q\,h^{2}_{t-1}, \tag{11}\\
	K_t &= W_K\!\bigl[h^{1}_{t}\;\bigl\|\;\phi(x^{f}_t)\bigr], \quad
	V_t = W_V\!\bigl[h^{1}_{t}\;\bigl\|\;\phi(x^{f}_t)\bigr], \tag{12}\\[2pt]
	\alpha_t &= \mathrm{sigmoid}\!\Bigl(\tfrac{1}{\sqrt{d}}\;Q_t K_t^{\!\top}\Bigr), \tag{13}\\[4pt]
	\mathbf{z}_t
	&= (1-\bar\beta_t)\!
	\odot\!
	\bigl(\alpha_t \odot V_t + \varepsilon V_t\bigr), \quad&
	\rho_t
	&= \bar\beta_t\!\odot h^{2}_{t-1}, \tag{14}\\[4pt]
	U_{t}
	&= \bigl[\,
	\mathbf{z}_t \;\bigl\|\;
	\rho_t 
	\bigr],\label{eq:relay} \quad&
	h_t^2 &= f_h^2(h_{t-1}^2, U{t}; \theta_h^2) \tag{15}.
\end{alignat}
This gate regulates how much newly acquired evidence updates the classifier state, which helps control the performance and human-likeness trade-off studied in this paper.

\subsection{Training}
We train the location policy with REINFORCE \cite{RAM,REINFORCE}. The reward $R$ is 1 if the final classification is correct and 0 otherwise. We subtract a baseline $b_t$ to reduce gradient variance:
\begin{equation}
	\mathcal{L}_{\mathrm{REINFORCE}} = -\sum_{t=1}^T (R - b_t) \log \pi(\ell_t \mid h^1_t; \theta), \label{eq6}\tag{16}
\end{equation}

where $\pi(\ell_t\mid h^1_t; \theta)$ is the policy distribution based on hidden state of the lower recurrent layer $H^1_t$. 
We also optimize a supervised classification loss on the classifier output:

\begin{equation}
	\mathcal{L}_{\mathrm{CE}}
	= -\sum_{t=1}^T \; y_T \;\log  y_t,
	\label{eqCE}\tag{17}
\end{equation}
where \( y_t\) is the predicted probability of class \(c\) and \(y_T\) is the one‐hot indicator of the true label.

The full objective is:
\begin{equation}
	\mathcal{L}
	= \mathcal{L}_{\mathrm{CE}}
	+ \mathcal{L}_{\mathrm{REINFORCE}}
	+ \lambda_{\text{cost}}\;\overline{\beta}_{b}
	+ \lambda_{1}\;\lVert \overline{\beta}_{b}\rVert_{1}
	+ \lambda_{H}\;\mathcal{H}_{\beta},
	\tag{18}
\end{equation}
where \(\overline{\beta}_{b}\) is the mean bottom-up gate value, and the entropy term
\begin{equation}
	\mathcal{H}_{\beta}
	= -\overline{\beta}_{t}\log(\overline{\beta}_{t} + \varepsilon)
	-\overline{\beta}_{b}\log(\overline{\beta}_{b} + \varepsilon)
	\label{eqEntropy}\tag{19}
\end{equation}

encourages non-trivial gating behavior. 
During training, we compute the prediction error $e_t$ from the supervised classification signal. This provides a stable uncertainty proxy for variance control. At test time, labels are unavailable, so we replace $e_t$ with a self-error signal derived from changes in the model predictions across glimpse steps. Training with self-error only reduces accuracy by about 5\%, so we use label-based error for training stability and use self-error at inference.

\subsection{Scanpath Similarity Metrics and Center-Debiased GCS}
\label{sec:ss-metric}

We report DTW, ScanMatch, NSS, and AUC \cite{DTW,ScanMatch,NSS,AUC}. 
To reduce center-bias inflation, we use the center-debiased Gaze Consistency Score (GCS) metric \cite{DebiasCentralFixation}. 
GCS normalizes each metric using a human upper reference and a corner-fixation lower reference, then subtracts an always-center reference to remove center-bias inflation. 
It then aggregates the debiased metrics and adds a movement-consistency term to penalize policies that match only spatial center overlap. 
Full details are provided in Appendix~\ref{sec:metrics details}.

\section{Experiments}
\label{sec:experiments}

\subsection{Performance on Image Classification Benchmarks}
\label{sec:imagenet}

We first evaluate \textbf{EVA} on CIFAR-10 and ImageNet-100. These datasets allow us to jointly assess classification accuracy, parameter efficiency, and gaze alignment under consistent training settings. Baselines include convolutional models (ResNet18, MobileNetV3), a transformer (ViT-tiny), and hard-attention models (RAM, DRAM, MRAM, Saccader). All models share the same glimpse size, number of steps, and optimizer hyperparameters to ensure a fair comparison. Details of experiments are described in Appendix~\ref{sec:model_details}

\begin{table}[!t]
	\centering
	\caption{Mechanistic analysis of the accuracy--alignment trade-off on CIFAR-10. The table is organized into blocks separated by horizontal rules. \textbf{Bold} denotes the best value within each block for a given metric. \revB{Light blue} marks two EVA variants. EVA-Mobile uses a pretrained MobileNetV3 backbone.}
	\label{tab:tab1}
	\scriptsize
	\setlength{\tabcolsep}{2pt}
	\renewcommand{\arraystretch}{1.05}
	\resizebox{1.0\columnwidth}{!}{%
		\begin{tabular}{lccccccccc}
			\toprule
			\textbf{Model} & \textbf{Params}   & \textbf{Time} & \textbf{FLOPs}
			& \textbf{Acc.}  & \textbf{DTW} & \textbf{SM} & \textbf{NSS}& \textbf{AUC}& \textbf{GCS}\\
			& \textbf{(M)} & \textbf{(ms/im)} & \textbf{(B)↓} & \textbf{(\%)}↑ & \textbf{↓} & \textbf{↑}  & \textbf{↑} & \textbf{↑}  & \textbf{↑}\\
			\midrule
			CNN (ResNet18)                           & 11.18& 0.89 ± 0.02 & 0.34& 78.00 & -& -& -& -& -\\
			\makecell[lt]{CNN\\(MobileNetV3)}        & 4.21 & 0.94 ± 0.08 & 0.03& \textbf{78.52} & -& -& -& -& -\\
			TinyViT                                  & 4.37 & 3.03 ± 0.22 & 2.66& 68.21 & -& -& -& -& -\\
			DeepGaze IIE                               & 0.2  & 0.82 ± 0.03 & -   & -  & 705.48 & 0.311 & 0.251 & 0.601 & 0.039\\
			Gazeformer                        & 0.27  & 0.04 ± 0.01 & -  & - & \textbf{669.22} & \textbf{0.346}& \textbf{0.757} & \textbf{0.689} & \textbf{0.068}\\
			\midrule
			Saccader                                 & 12.49& 3.29 ± 0.31 & 4.44& 77.80 & 928.44 & 0.276& 0.277& 0.665& -0.001\\
			RAM, 1scale                              & 0.65 & 2.21 ± 0.08 & 0.01& 62.27 & 1176.54& 0.241& 0.228& 0.654& -0.041\\
			RAM, 2scale                              & 0.68 & 3.22 ± 0.20 & 0.02& 61.55 & 1173.89& 0.258& 0.377& 0.684& -0.016\\
			DRAM, 1scale                             & 2.24 & 1.94 ± 0.10 & 1.74& 64.81 & 1069.77& 0.248& 0.224& 0.656& -0.021\\
			DRAM, 2scale                             & 2.25 & 3.91 ± 0.35 & 1.75& 62.17 & 837.14 & 0.302& 0.665& 0.679&  0.044\\
			MRAM, 1scale                             & 1.18 & 2.34 ± 0.13 & 0.01& 64.18 & 930.38 & 0.274& 0.318& 0.667&  0.026\\
			MRAM, 2scale                             & 1.21 & 3.53 ± 0.28 & 0.02& 58.24 & 945.31 & 0.263& 0.315& 0.674& 0.057\\
			
			\revB{EVA-Mobile} & 4.79 & 5.59 ± 0.49 &0.45&76.14&\textbf{825.11}&\textbf{0.322}&\textbf{0.786}& \textbf{0.701}&\textbf{0.080}\\
			
			\revB{EVA}                       & 2.97 & 3.06 ± 0.04 & 1.37& \textbf{79.77} & 856.29 & 0.316& 0.481& 0.692& 0.064\\
			\midrule
			\makecell[lt]{EVA\\(w/o CNN)}            & 1.93 & 3.12 ± 0.45 & 0.07& 62.41 & 800.30 & 0.327& 0.511& \textbf{0.703}& 0.057\\
			\makecell[lt]{EVA\\(CNN only)}           & 2.22 & 3.91 ± 0.09 & 1.32& 69.99 & 1019.85& 0.253& 0.264& 0.663&  -0.023\\
			\makecell[lt]{EVA\\(w/o gate)}           & 2.48 & 2.96 ± 0.14 & 1.32& \textbf{78.96} & 863.70 & 0.308& 0.391& 0.686&  0.026\\
			\makecell[lt]{EVA\\(gate only)}          & 1.67 & 2.92 ± 0.06 & 0.07& 55.61 & 797.79 & 0.318& \textbf{0.702}& 0.681&  0.051\\
			\makecell[lt]{EVA\\(w/o error)}          & 2.97 & 3.32 ± 0.11 & 1.37& 75.14 & 894.17 & 0.300& 0.386& 0.691&  0.032\\
			\makecell[lt]{EVA\\(error only)}         & 1.47 & 3.24 ± 0.42 & 0.02& 63.36 & 824.51 & 0.321& 0.483& 0.702&  0.053\\
			\makecell[lt]{EVA\\(train self-error)}   & 2.97 & 3.24 ± 0.35 & 1.37& 73.78 & \textbf{792.89} & \textbf{0.330}& 0.608& 0.700&  \textbf{0.065}\\
			\bottomrule
	\end{tabular}}
\end{table}
As summarized in Table~\ref{tab:tab1}, EVA achieves the strongest classification accuracy among the hard-attention baselines on CIFAR-10 under the same glimpse budget and training protocol. At the same time, EVA attains competitive scanpath alignment as measured by DTW, SM, NSS, AUC, and the center-debiased GCS. EVA-Mobile illustrates a clear accuracy--alignment trade-off. It slightly reduces accuracy while improving scanpath alignment, particularly under saliency-based metrics and GCS. This behavior is consistent with the alignment tax studied in this paper, where stronger representations can improve accuracy while weakening the human-likeness of sequential evidence acquisition.
The table also includes gaze-supervised scanpath predictors such as DeepGaze and Gazeformer. These models are trained to fit human gaze and therefore perform strongly on scanpath similarity metrics, but they do not address classification accuracy under limited glimpses. EVA differs in that it is trained only with classification labels and learns its attention policy from classification reward. The comparison highlights a complementary point. Task-driven hard-attention agents can produce human-like scanpaths without gaze supervision while remaining competitive on classification accuracy. Additional qualitative results and stability analyses are provided in the appendix.

\paragraph{Ablation studies.}
Table~\ref{tab:tab1} reports controlled ablations that isolate the roles of the three EVA components. Removing the CNN feature extractor severely degrades accuracy, confirming that strong foveal representation is the primary driver of task performance. In contrast, removing variance control or gating affects scanpath alignment more than accuracy. The CNN-only variant improves accuracy but yields poor GCS, consistent with the alignment tax in which stronger representations reduce the need for human-like sequential sampling. Adding variance control and gating restores human-aligned trajectories and improves GCS with limited impact on accuracy. Using only variance control or only gating can increase alignment metrics but fails to recover accuracy, indicating that these mechanisms regulate evidence acquisition rather than replace representation learning. Training variance control using self-error during training reduces accuracy by about 5\% compared to label-based error, suggesting that a stable training signal is important for learning the exploration dynamics. EVA-Mobile further shows that a stronger pretrained backbone can improve alignment under a modest accuracy cost. Overall, the components act synergistically. The CNN supports recognition, while variance control and gating regulate exploration and information flow to reduce the alignment tax.

\begin{figure}[t]
	\centering
	\setlength{\tabcolsep}{1pt}
	\renewcommand{\arraystretch}{0.0}
	\begin{tabular}{c@{\hspace{3pt}}cccccc}
		& \textbf{Gazeformer}
		& \textbf{Saccader}
		& \textbf{DRAM}
		& \textbf{EVA (ours)}
		& \textbf{EVA-Mobile (ours)} \\[2pt]
		\textbf{Dog} &
		\includegraphics[width=0.17\textwidth]{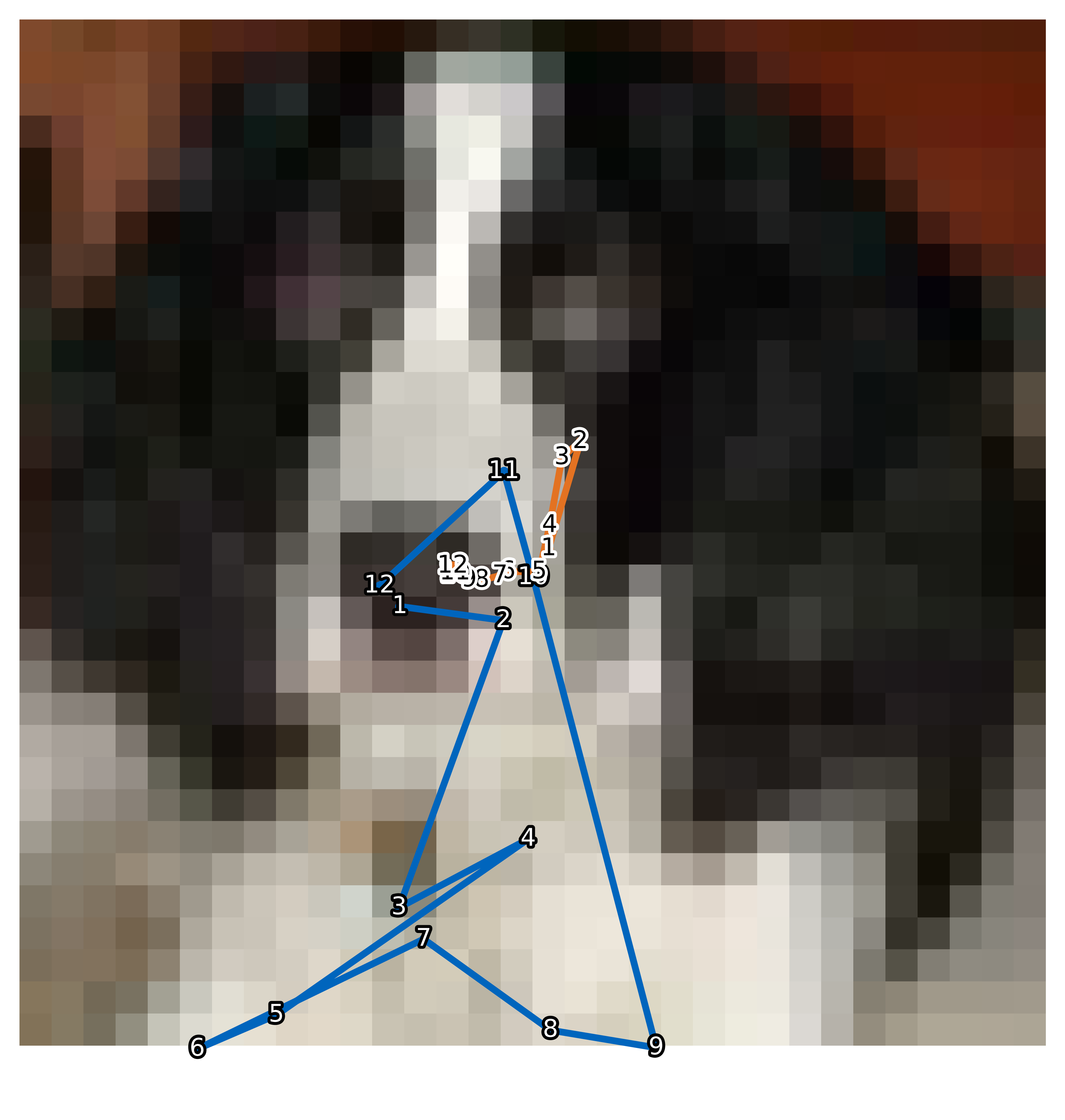} &
		\includegraphics[width=0.17\textwidth]{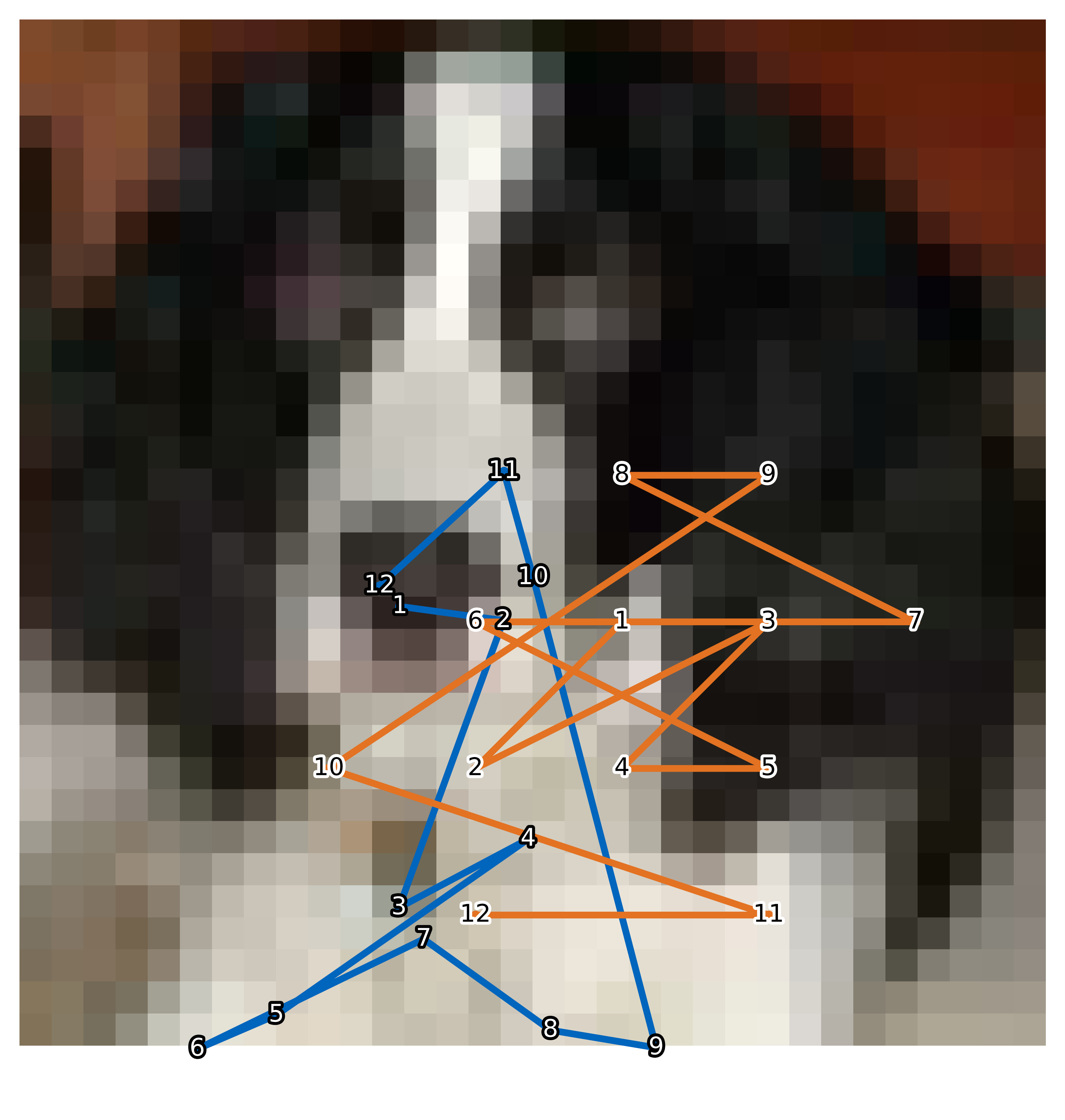} &
		\includegraphics[width=0.17\textwidth]{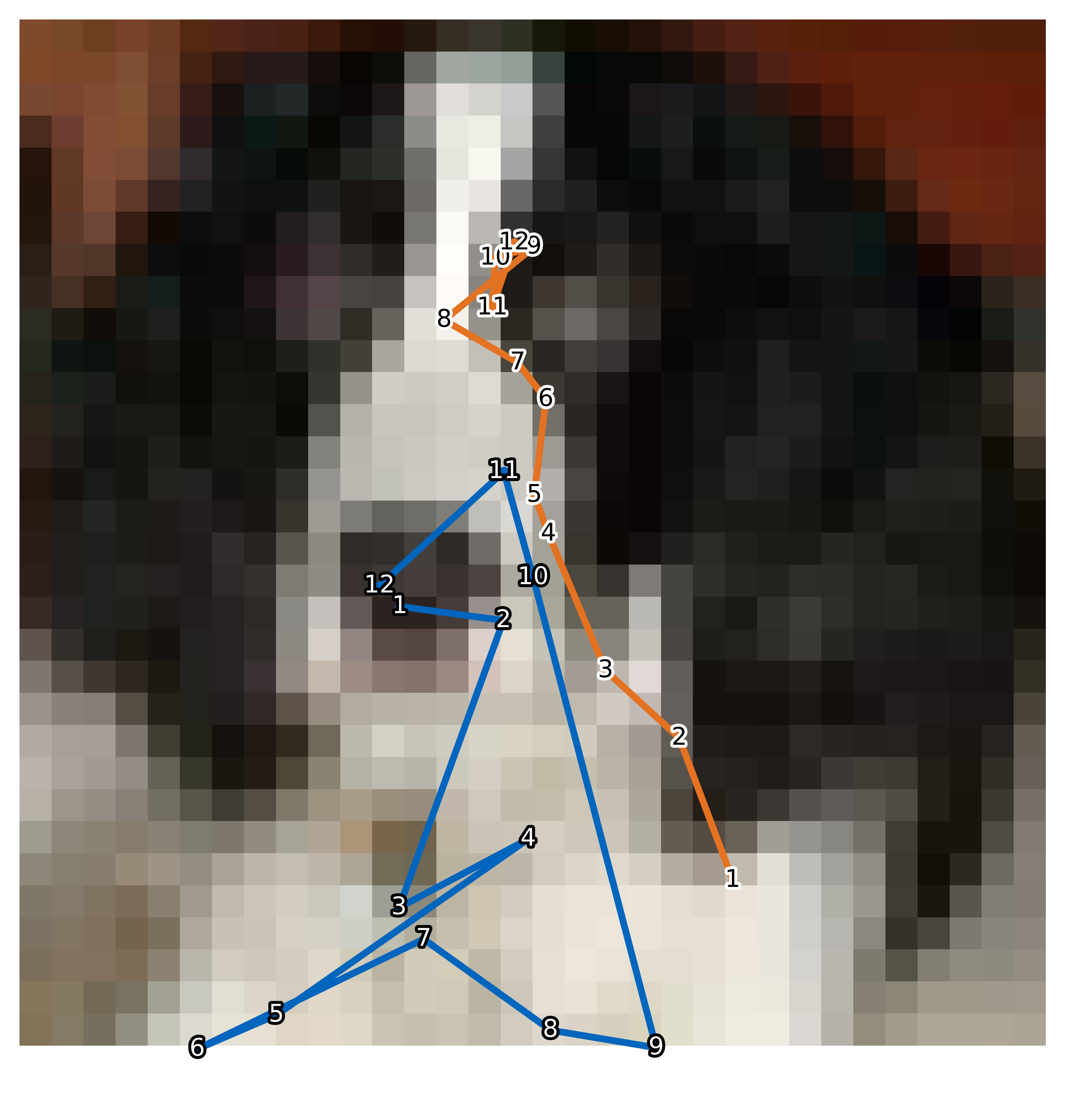} &
		\includegraphics[width=0.17\textwidth]{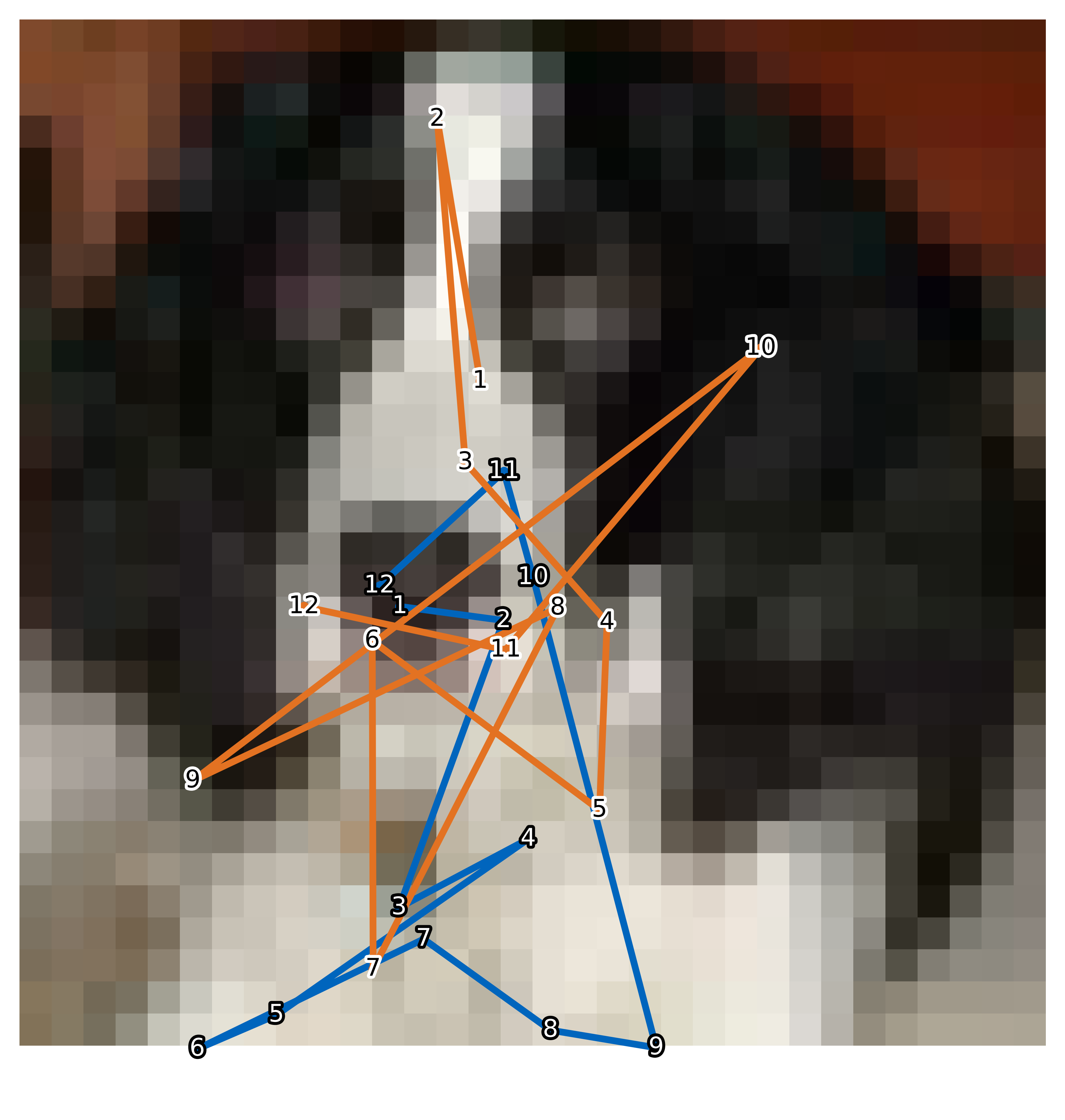} &
		\includegraphics[width=0.17\textwidth]{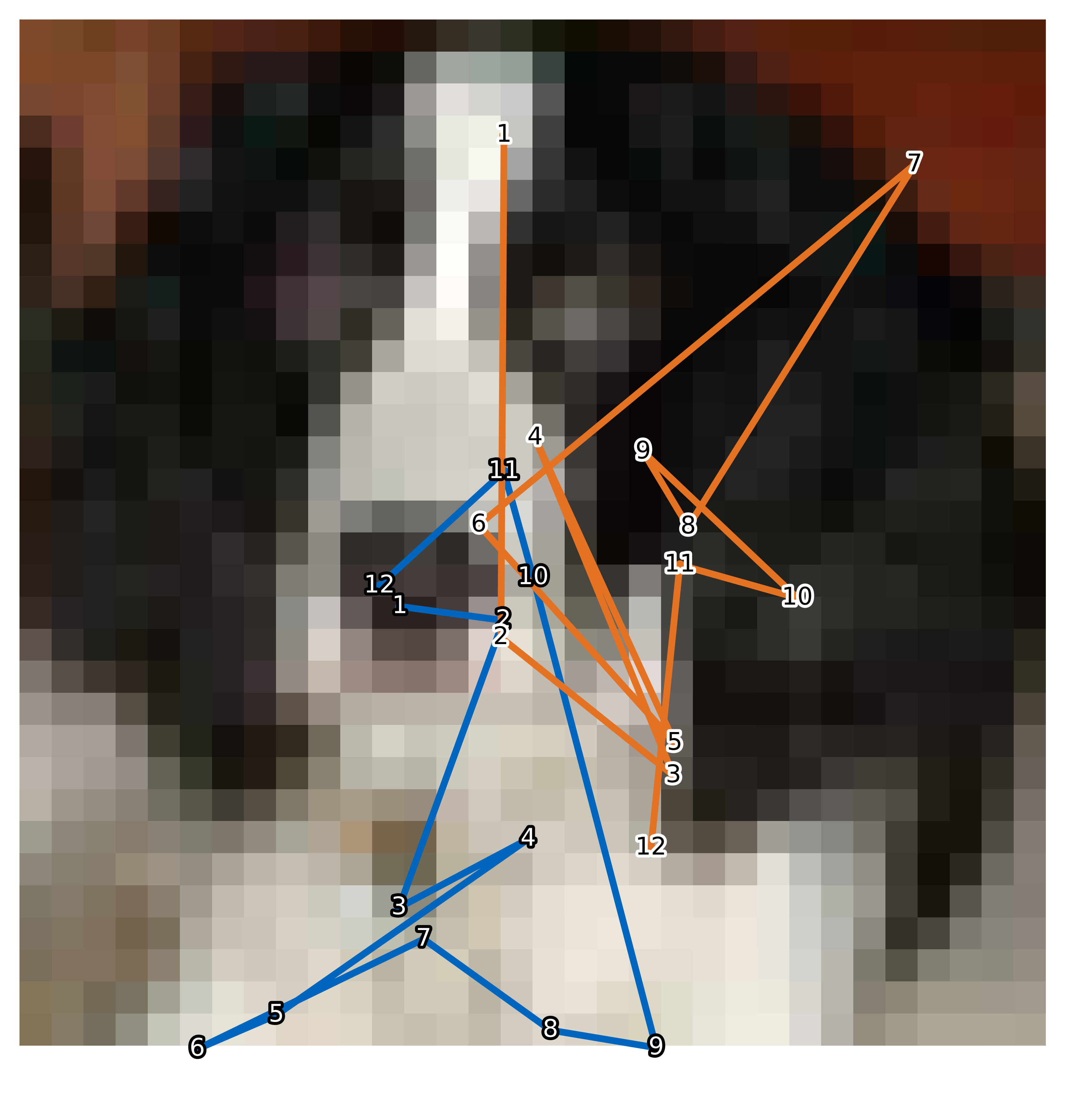} \\[2pt]
		\textbf{Bird} &
		\includegraphics[width=0.17\textwidth]{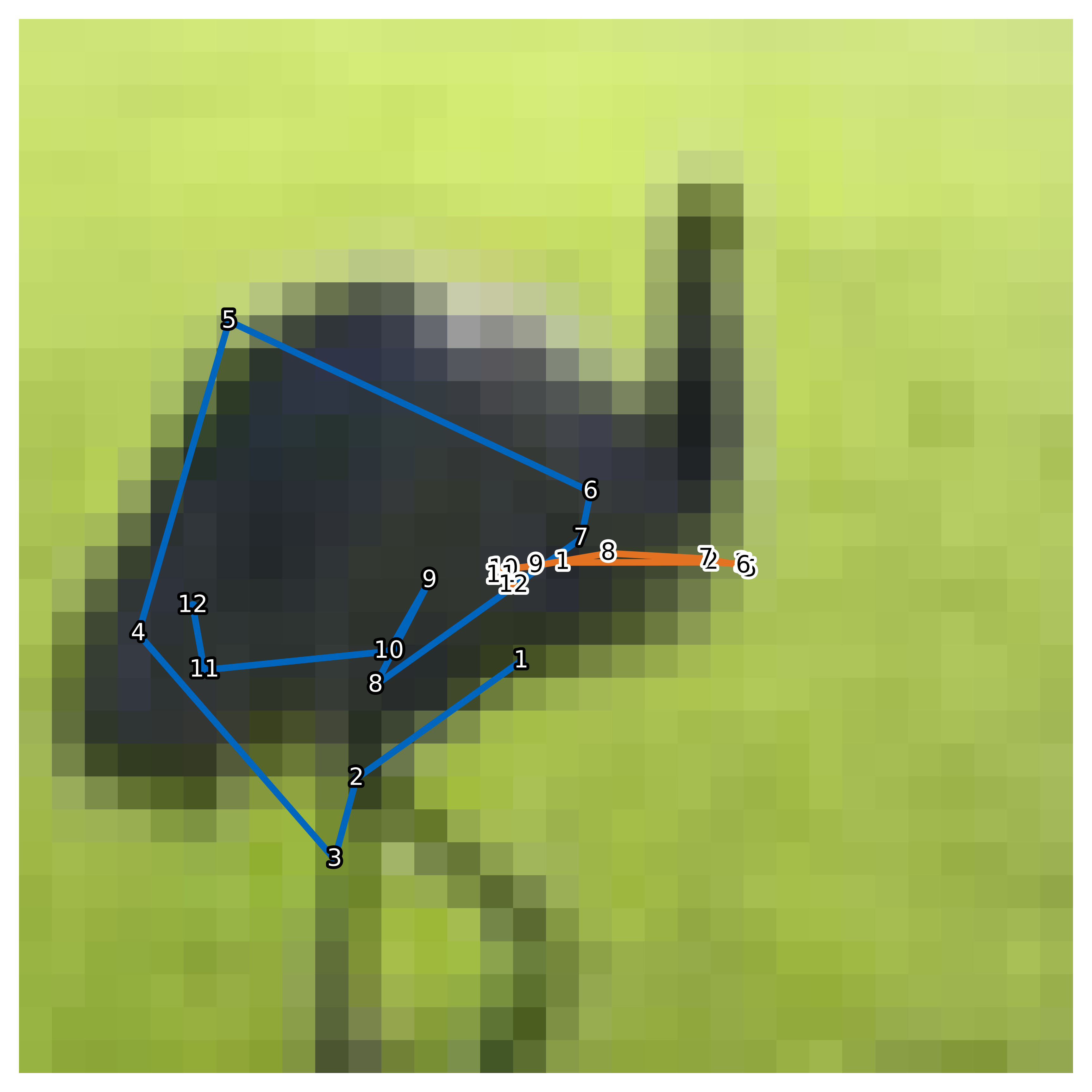} &
		\includegraphics[width=0.17\textwidth]{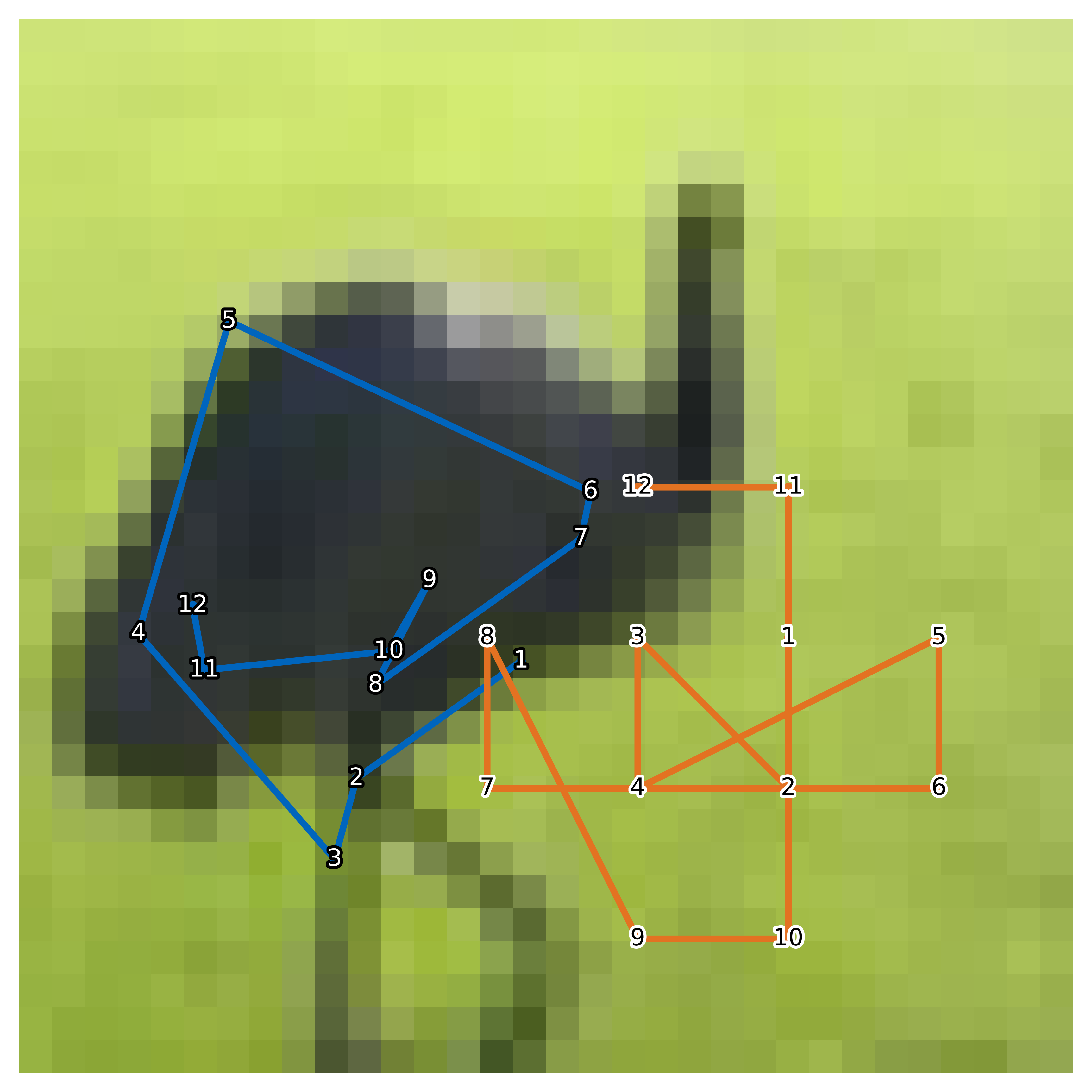} &
		\includegraphics[width=0.17\textwidth]{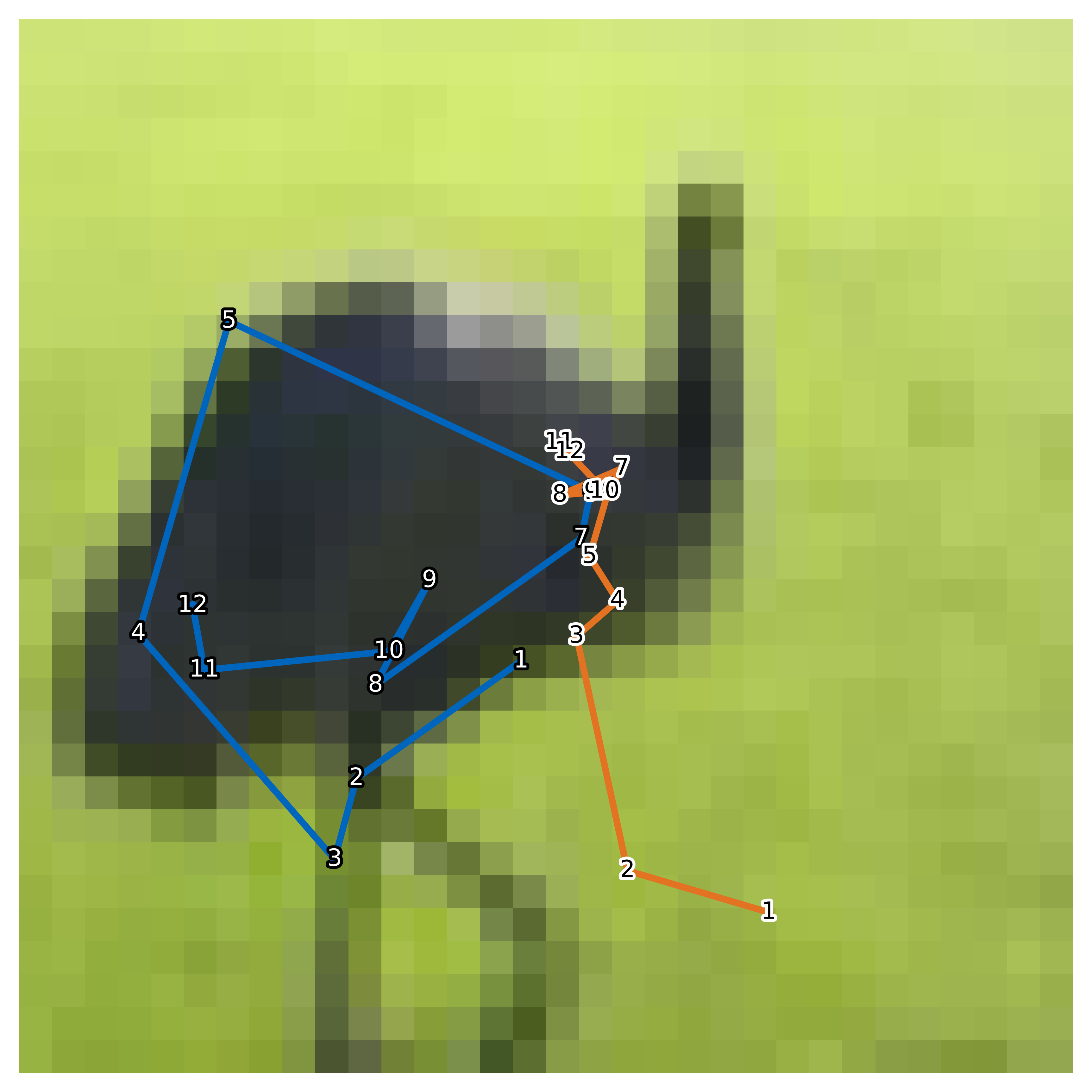} &
		\includegraphics[width=0.17\textwidth]{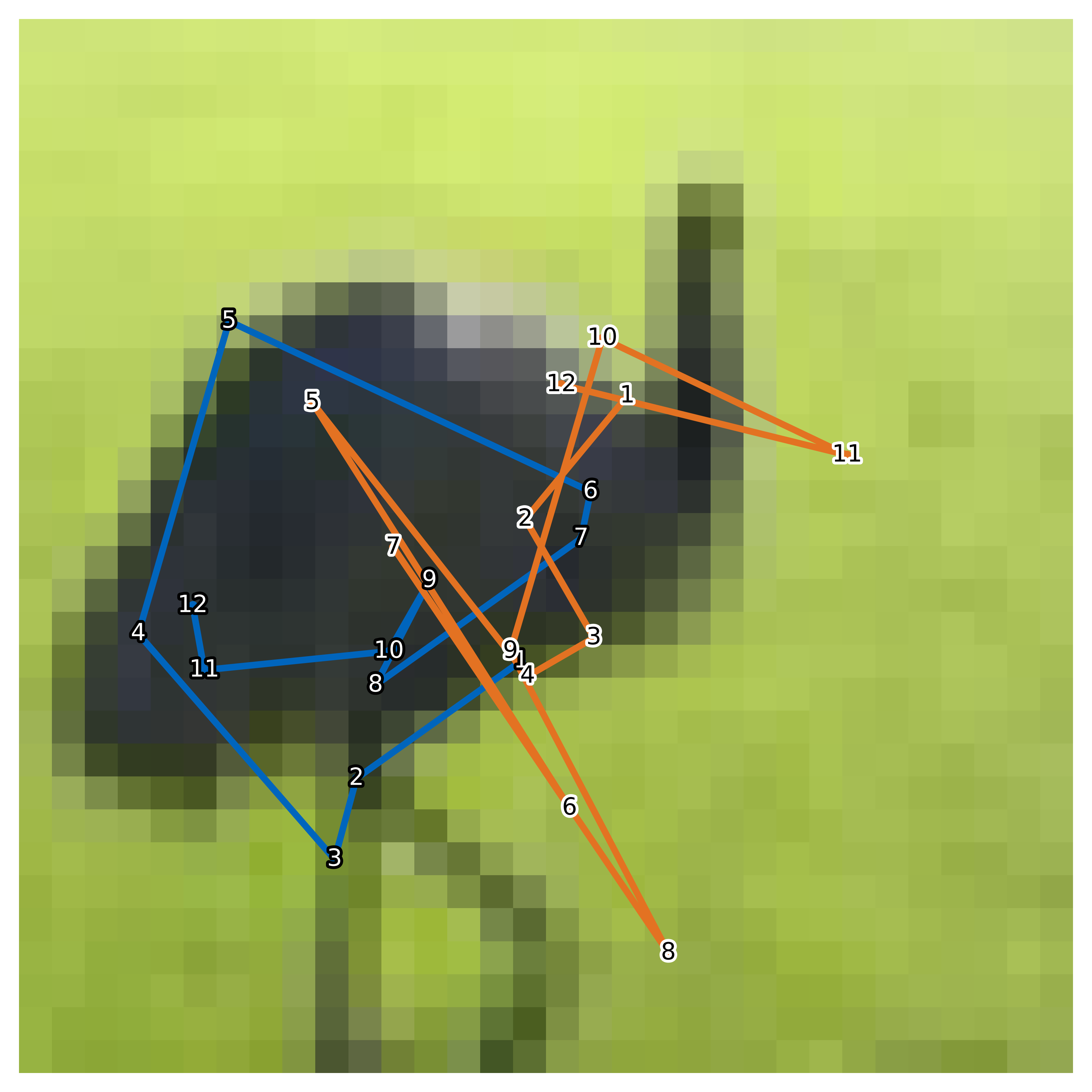} &
		\includegraphics[width=0.17\textwidth]{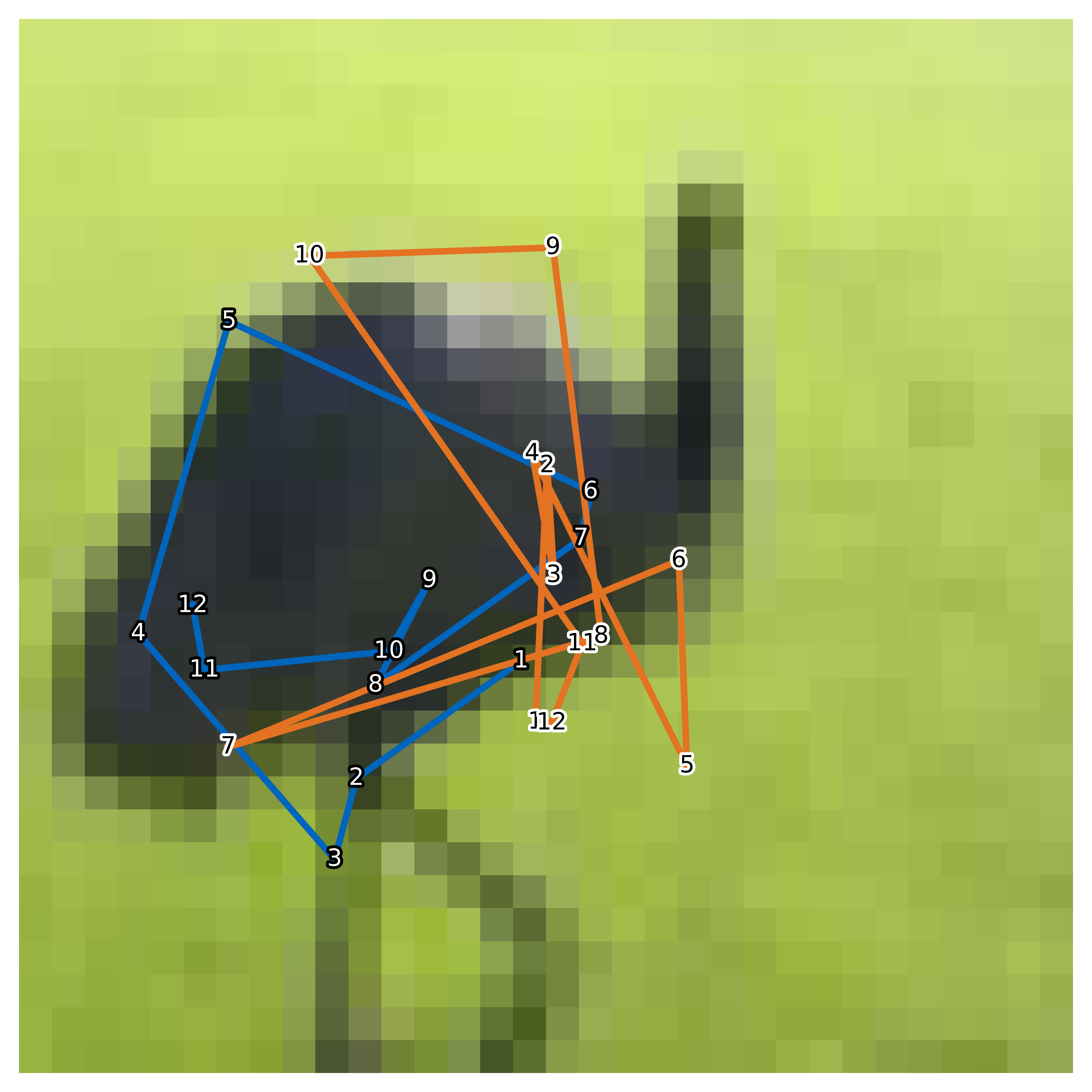} \\[2pt]
		\textbf{Car} &
		\includegraphics[width=0.17\textwidth]{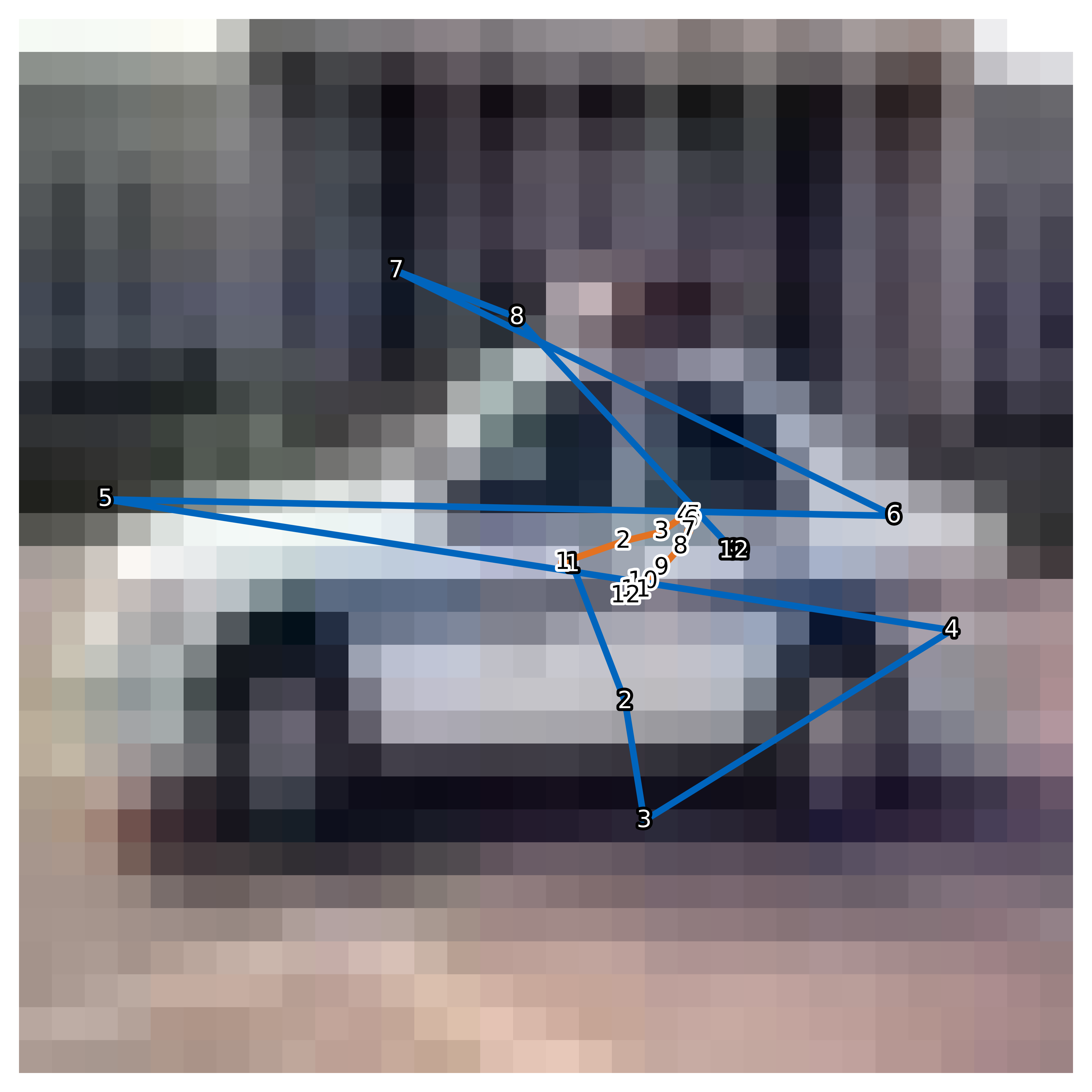} &
		\includegraphics[width=0.17\textwidth]{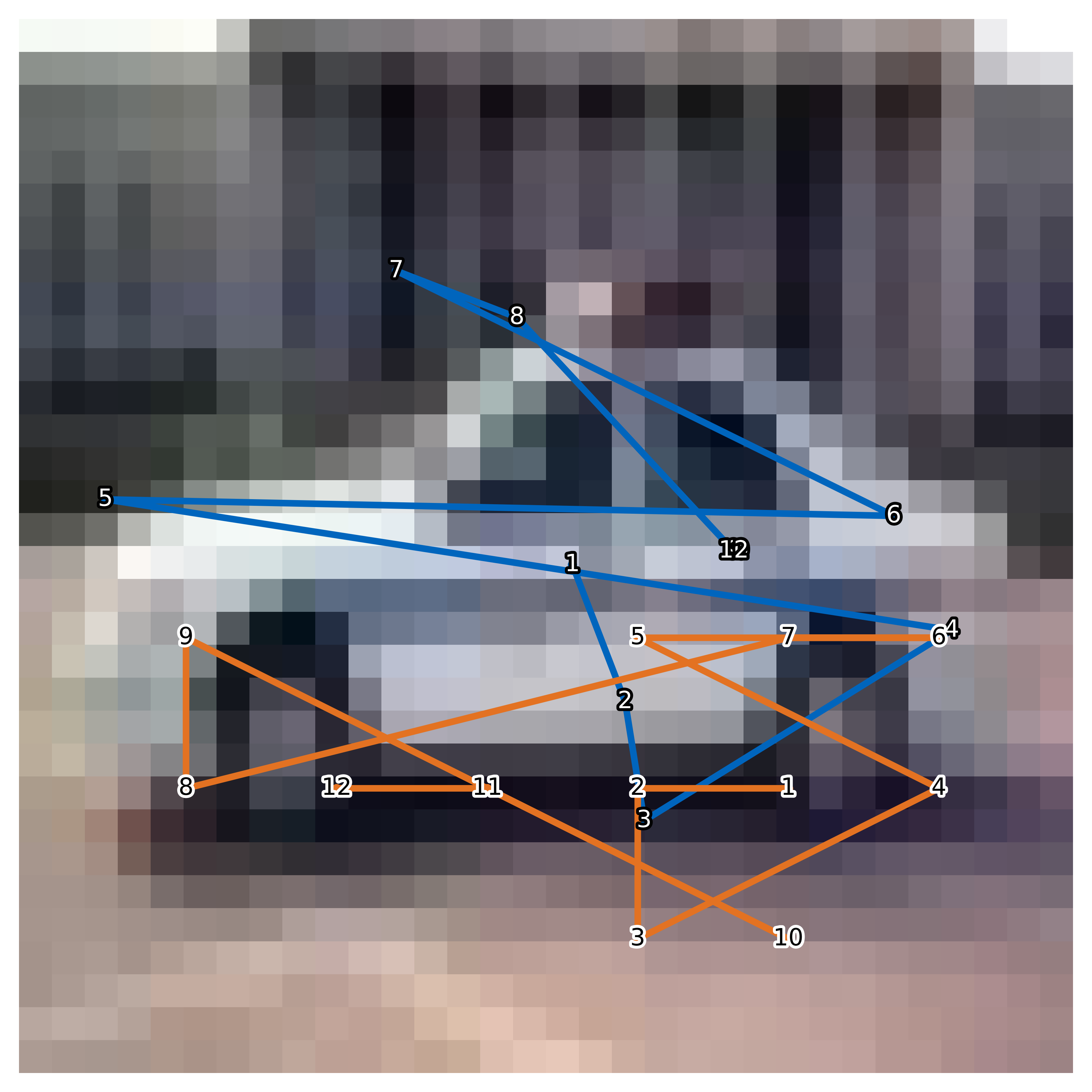} &
		\includegraphics[width=0.17\textwidth]{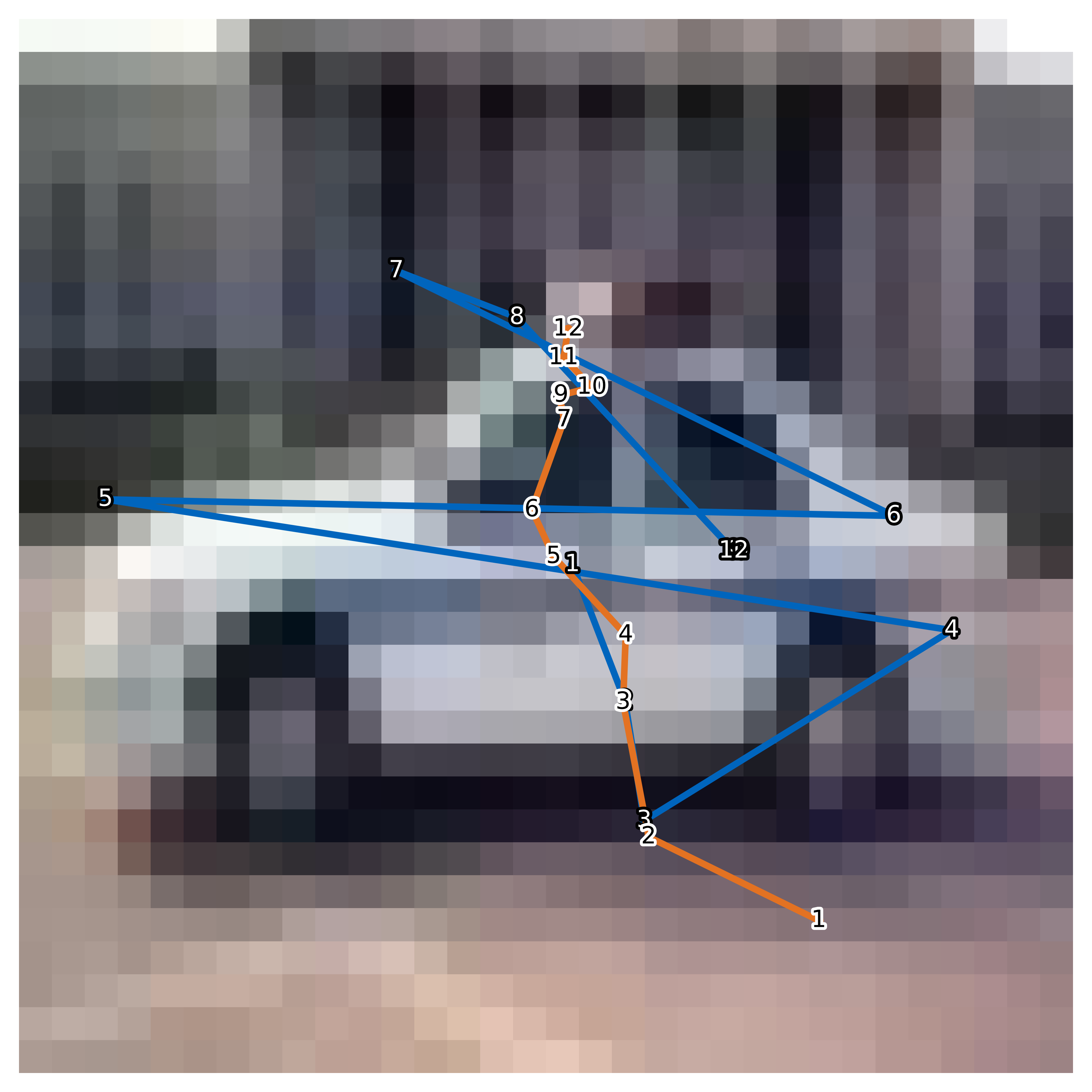} &
		\includegraphics[width=0.17\textwidth]{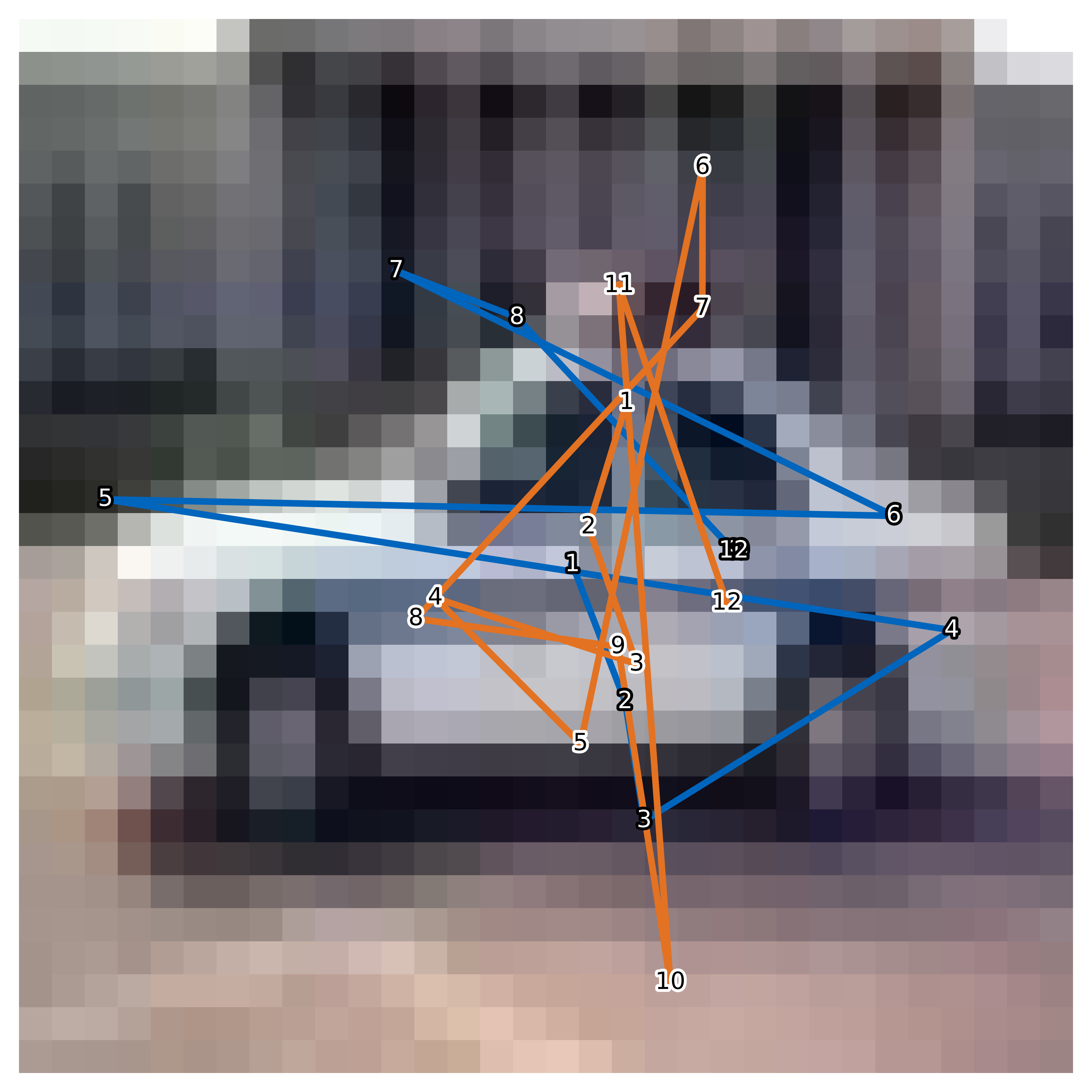} &
		\includegraphics[width=0.17\textwidth]{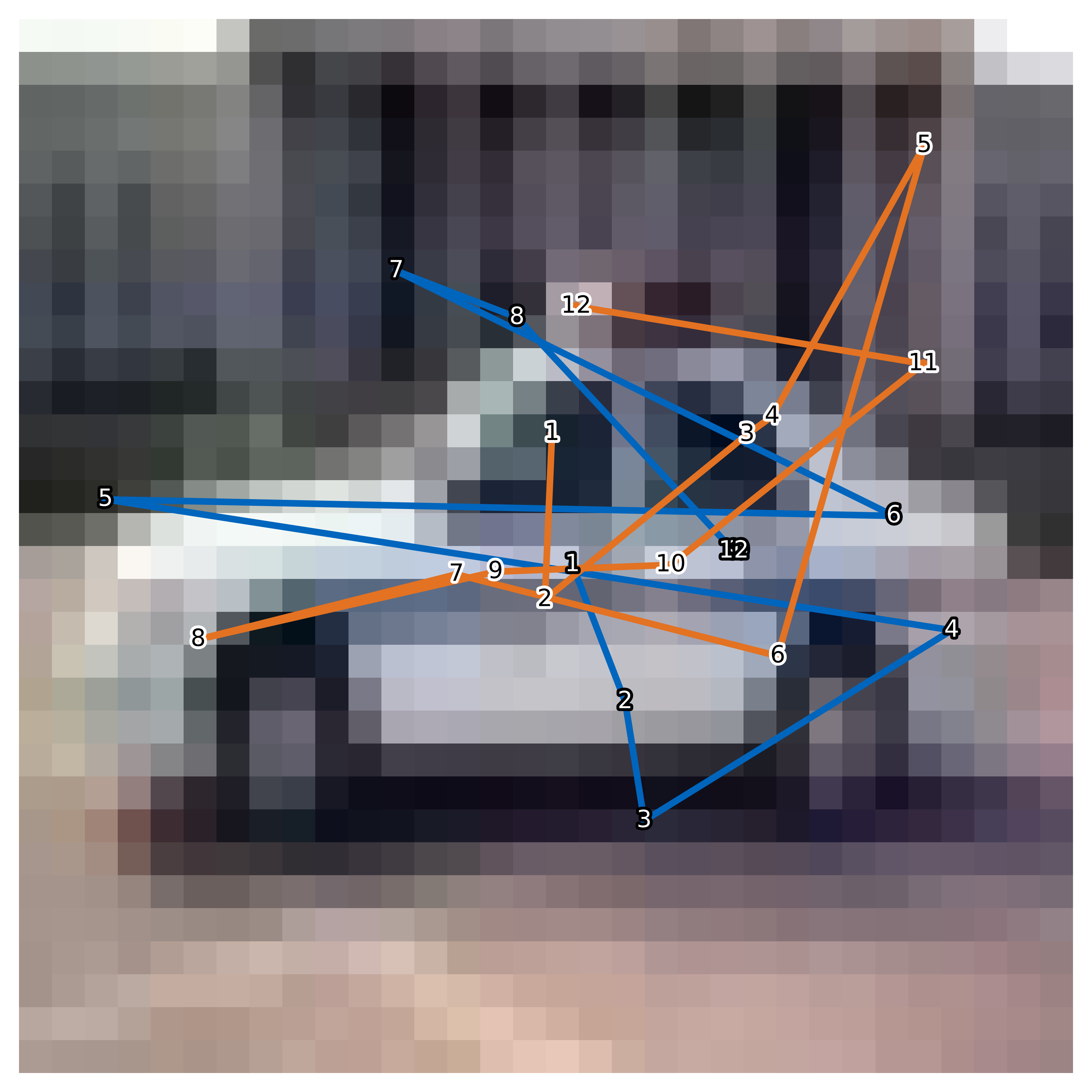} \\
	\end{tabular}
	\caption{Qualitative comparison of scanpaths on CIFAR-10. Columns show different models and rows show example images. Orange denotes the model scanpath and blue denotes the human scanpath.}
	\label{fig:cifar_scanpaths}
\end{figure}

\begin{figure}[htbp]
	\centerline{\includegraphics[scale=0.47]{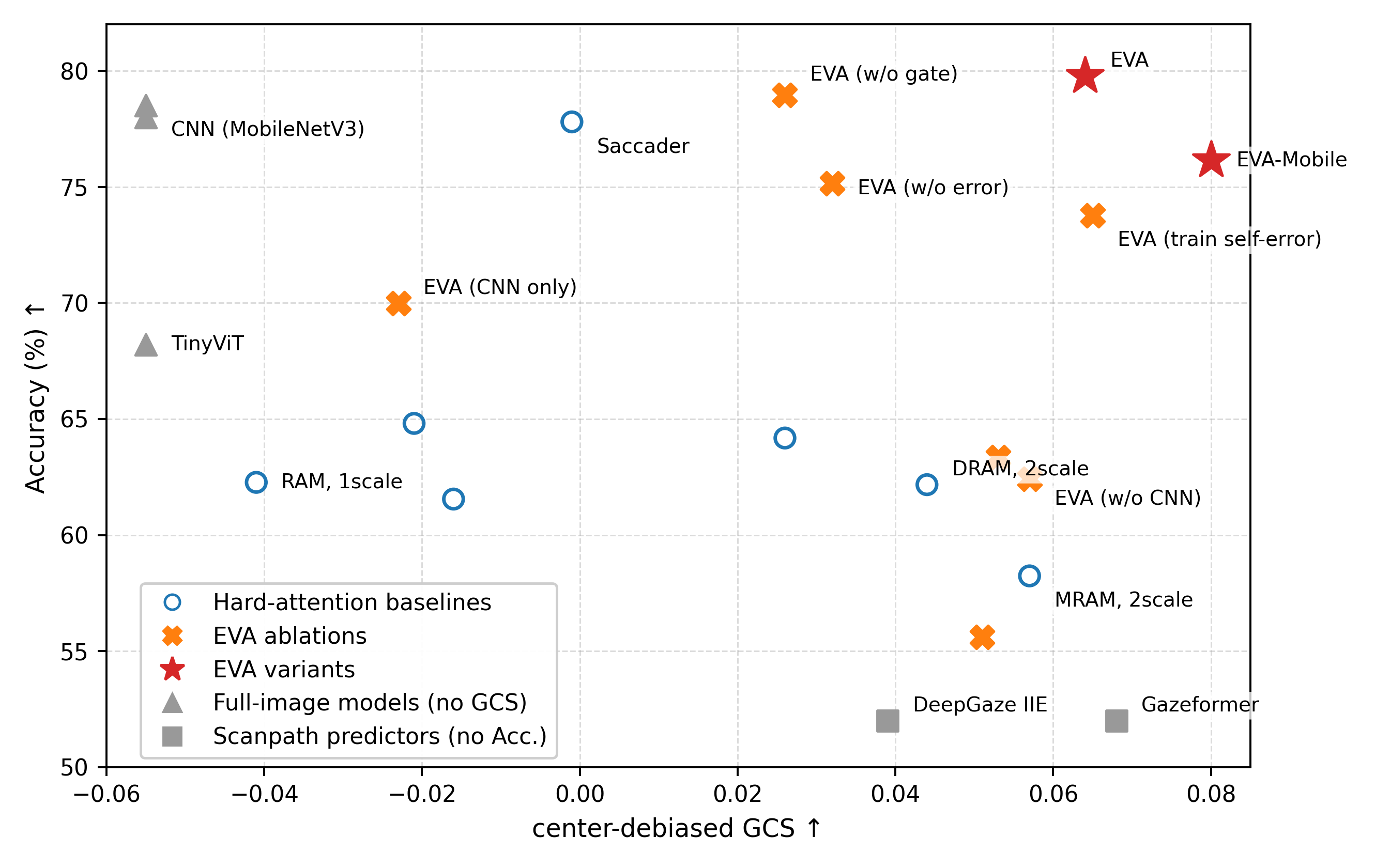}}
	\caption{Accuracy--alignment trade-off on CIFAR-10. The x-axis is center-debiased GCS and the y-axis is classification accuracy.}
	\label{fig13}
\end{figure}

\subsection{Emergent Human-like Scanpaths on CIFAR-10}

To assess whether EVA produces human-aligned scanpaths beyond the scalar metrics in Table~\ref{tab:tab1}, we visualize fixation trajectories on CIFAR-10. Figure~\ref{fig:cifar_scanpaths} compares model scanpaths to human scanpaths from Gaze-CIFAR-10 for representative models. For hard-attention models, the fixation sequence is an internal variable learned from classification reward and used only to support image recognition. For gaze-supervised predictors such as Gazeformer, scanpaths are directly optimized to match human gaze. Despite this difference in supervision, EVA often allocates fixations to semantically informative regions, such as faces, wheels, and object centers, and shows substantial overlap with human trajectories. These qualitative results provide a process-level form of interpretability. The model exposes a discrete sequence of evidence requests that can be inspected and compared to human visual sampling behavior.

Figure~\ref{fig13} summarizes the accuracy and scanpath alignment of hard-attention models on CIFAR-10. We use the center-debiased GCS as the primary scalar summary of scanpath similarity. EVA achieves a favorable accuracy--alignment trade-off among hard-attention baselines. EVA-Mobile further improves scanpath alignment while sacrificing some accuracy. This pattern is consistent with the alignment tax. Stronger representations can improve accuracy while weakening the need for human-like sequential sampling, and the variance control and gating mechanisms help recover human-aligned scanpaths without catastrophic accuracy loss.

\section{Scalability Check to Diverse Datasets and Tasks}
\label{sec:model-details}

\subsection{High-Resolution Image Classification on ImageNet-100}
\begin{figure*}[t]
	\centering
	\begin{minipage}[t]{0.90\textwidth}
		\vspace{0pt}
		\centering
		\scriptsize
		\setlength{\tabcolsep}{4pt}
		\renewcommand{\arraystretch}{1.05}
		\begin{tabular}{lccc}
			\toprule
			\textbf{Model} & \textbf{FLOPs (B)} & \textbf{Top-1} & \textbf{Top-5} \\
			\midrule
			MobileNet (scratch) & 2.2 & 47.64 & 76.12 \\
			MobileNet (pre.)    & \textbf{2.2} & \textbf{80.36} & \textbf{96.5}  \\
			\midrule
			Saccader-Mobile (pre.) & 37.27 & \textbf{75.9} & \textbf{94.5} \\
			RAM   & \textbf{0.35} & 11.26 & 32.78 \\
			DRAM-Mobile (pre.) & 8.83 & 19.00 & 43.12 \\
			DRAM  & 1.06 & 21.44 & 47.3  \\
			MRAM  & 0.35 & 12.88 & 34.76 \\
			\midrule
			EVA-Mobile (scratch) & 8.87 & 42.62 & 68.24 \\
			EVA-Mobile (pre.)    & 8.87 & \textbf{71.92} & \textbf{91.92} \\
			EVA                  & \textbf{6.6} & 65.86 & 80.24 \\
			\bottomrule
		\end{tabular}
	\end{minipage}\hfill
	\begin{minipage}[t]{0.37\textwidth}
		\vspace{0pt}
		\centering
		\includegraphics[width=\linewidth]{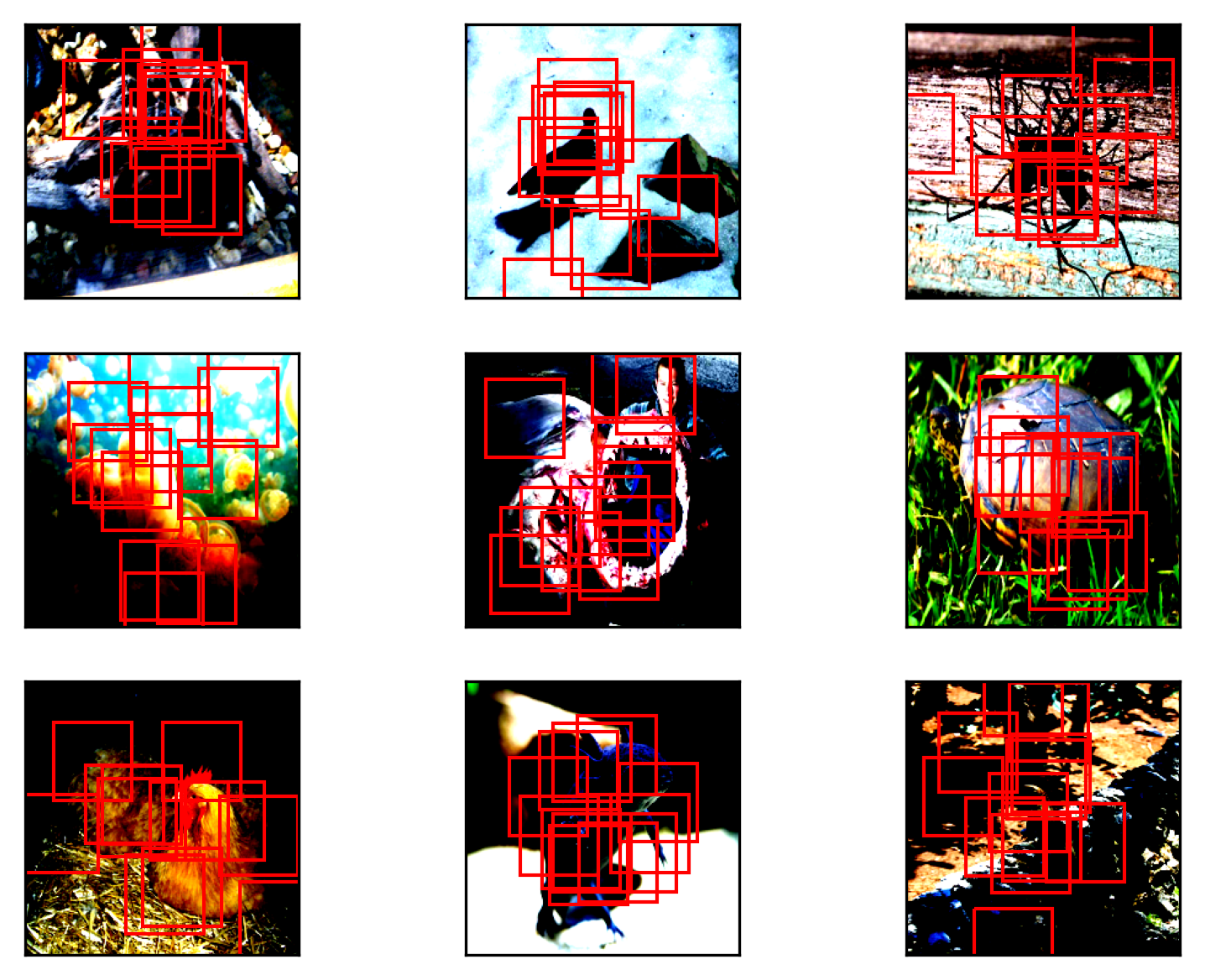}
	\end{minipage}
	
	\caption{ImageNet-100 scalability. Top: ImageNet-100 accuracy and FLOPs. Pre. indicates initialization from ImageNet-1K pretrained weights. Bottom: an example scanpath from EVA where the red rectangle indicates the foveal crop. }
	\label{fig:imagenet_combo}
\end{figure*}

We evaluate EVA on ImageNet-100 to test scalability to high-resolution natural images. Table~\ref{fig:imagenet_combo} reports accuracy and compute. EVA-Mobile processes only a small number of high-resolution foveal glimpses with the CNN backbone, while the low-resolution peripheral view is encoded with a lightweight module. In contrast, Saccader first applies a CNN densely over the image to produce logits at many spatial locations, then selects and aggregates locations for prediction. This design reduces the number of locations used for the final readout, but it still requires dense evaluation across the image. As a result, Saccader-Mobile achieves higher top-1 and top-5 accuracy than EVA-Mobile, but at substantially higher FLOPs. EVA enforces a stricter foveation constraint and uses less visual information than dense-logit hard-attention baselines, which makes the setting closer to active sampling under limited observation.

\subsection{Cross-task Scanpath Evaluation on COCO-Search18}
We evaluate whether a policy learned from classification reward can transfer as a human-aligned evidence acquisition strategy in a different behavioral regime. We train EVA on COCO image classification using labels only. Scanpaths are not supervised. We then compare EVA scanpaths to human search-conditioned scanpaths on COCO-Search18 without using COCO-Search18 gaze annotations or search supervision, and without finetuning. Table~\ref{tab:tab11} reports standard scanpath metrics. The reported accuracy on COCO-Search18 is the image classification accuracy over the original COCO categories restricted to the COCO-Search18 image subset. It is included for reference and is not a measure of search-task performance. Gazeformer is included as a gaze-supervised reference. EVA-Mobile with pretrained features improves scanpath alignment compared to other task-driven hard-attention baselines, while operating under the strict foveation constraint of using only glimpsed patches as high-resolution input. This evaluation complements the ImageNet-100 results by testing generalization of the learned sampling dynamics beyond the training task.In 

\begin{table}[!t]
	\centering
	\caption{COCO-Search18: scanpath metrics under gaze-supervision-free, cross-task evaluation. EVA is trained on COCO classification labels only and evaluated against human search-conditioned scanpaths without finetuning.}
	\label{tab:tab11}
	\begin{tabular}{lcccccc}
		\toprule
		\textbf{Model}  & \textbf{COCO}  & \textbf{COCO-Search18} & \textbf{DTW} & \textbf{SM} & \textbf{NSS} & \textbf{AUC}\\
		& \textbf{Acc.\%} & \textbf{Acc.\%} & \textbf{↓} & \textbf{↑}  & \textbf{↑} & \textbf{↑} \\
		\midrule
		CNN MobileNet (pre.)      & \textbf{58.82} & \textbf{27.5}  & -     & -     & -      & - \\
		Gazeformer                & -     & -     & \textbf{168.39}& \textbf{0.571} & \textbf{1.961}  & \textbf{0.8} \\
		\midrule
		Saccader-Mobile (pre.)    & \textbf{57.1}  & \textbf{16.7}  & 333   & 0.242 & \textbf{0.361}  & 0.658 \\
		RAM                       & 34.81 & 12.83 & 500.01& 0.072 & -0.132 & 0.585 \\
		DRAM                      & 43.32 & 14.34 & 530.78& 0.077 & -0.124 & 0.587 \\
		MRAM                      & 35.86 & 14.17 & 624.57& 0.015 & -0.09  & 0.605 \\
		EVA-Mobile (scratch)      & 45.81 & 14.79 & 513.48& 0.101 & -0.074 & 0.593 \\
		EVA-Mobile (pre.)         & 55.82 & 16.63 & \textbf{280.29}& \textbf{0.313} & 0.307  & \textbf{0.714} \\
		\bottomrule
	\end{tabular}
\end{table}

\subsection{PCA Visualization of Hidden State Dynamics}
We visualize EVA's internal dynamics by applying Principal Component Analysis (PCA) to the recurrent states from the lower layer $h_t^1$ and the upper layer $h_t^2$ over glimpse steps $t=1,\dots,T$. For each layer, we collect hidden states on the CIFAR-10 test set and fit PCA on the pooled states. We then project each hidden state into the first two principal components. We plot class-wise mean trajectories obtained by averaging hidden states over samples from the same class at each step.

\begin{figure}[t]
	\centering
	\setlength{\tabcolsep}{2pt}
	\begin{subfigure}[b]{0.34\columnwidth}
		\centering
		\includegraphics[width=\linewidth]{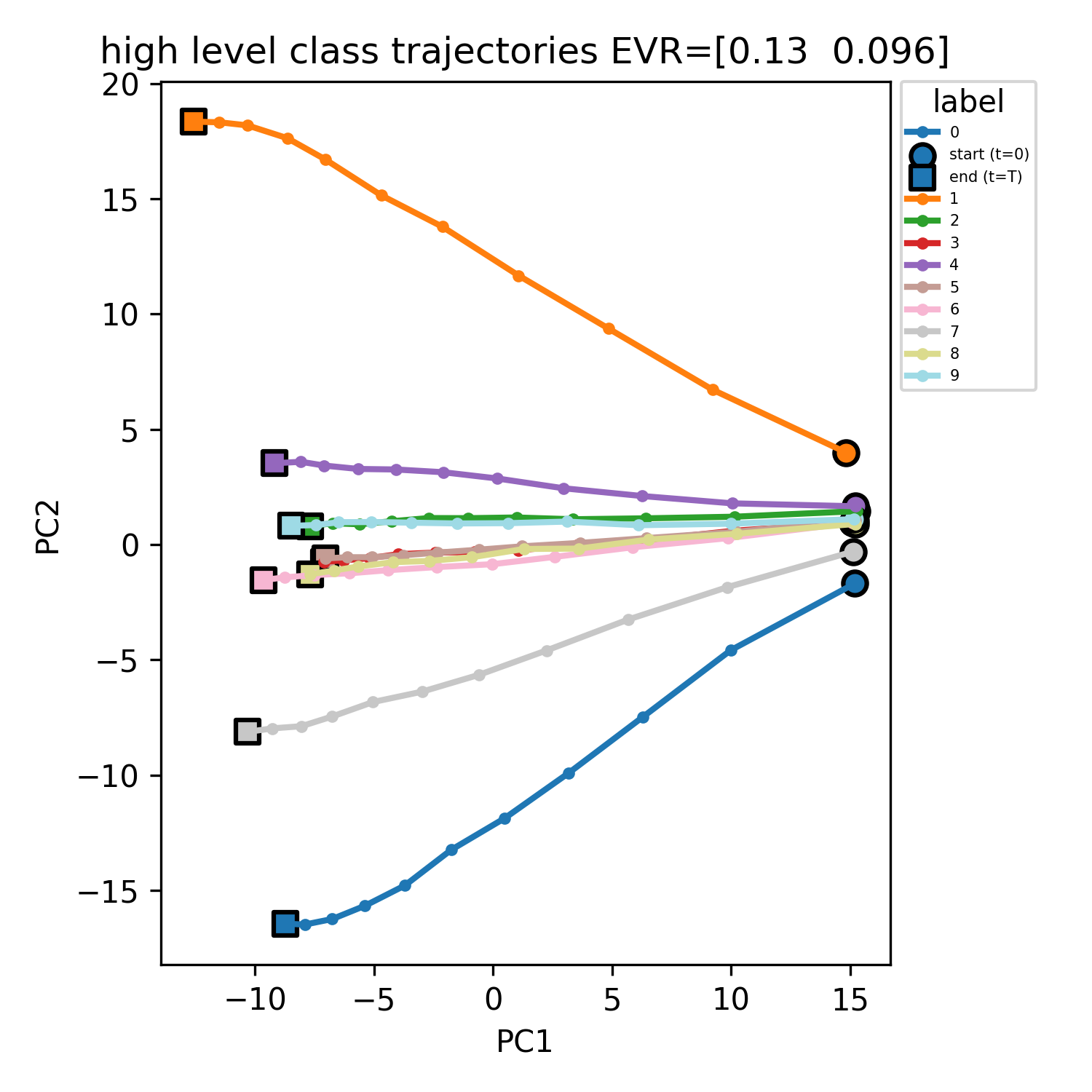}
		\caption{Upper layer $h_t^2$}
		\label{fig:pca_high_mean}
	\end{subfigure}
	\begin{subfigure}[b]{0.34\columnwidth}
		\centering
		\includegraphics[width=\linewidth]{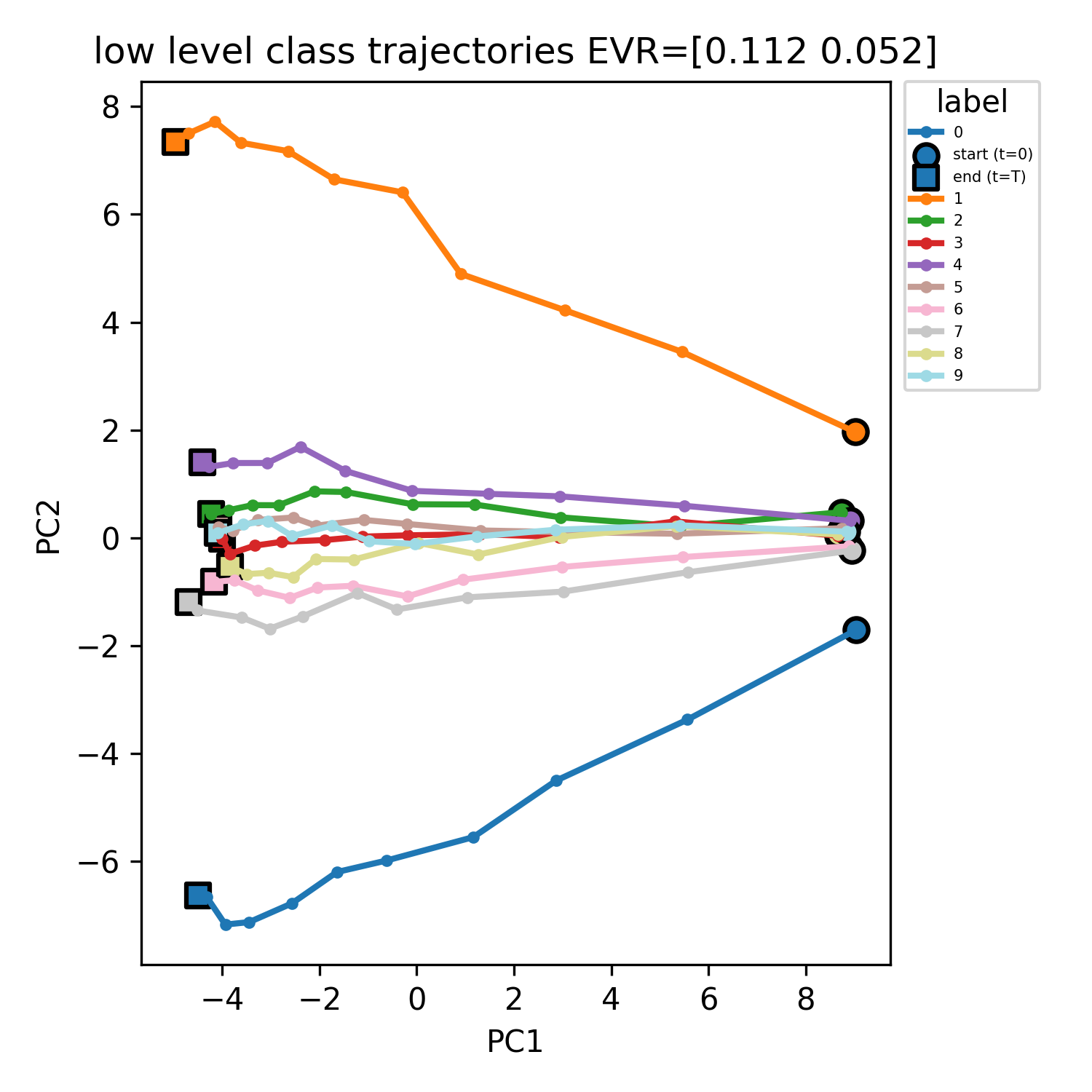}
		\caption{Lower layer $h_t^1$}
		\label{fig:pca_low_mean}
	\end{subfigure}
	\caption{PCA of EVA recurrent states on CIFAR-10 using class-wise mean trajectories over glimpse steps. The upper-layer state separates classes more clearly than the lower-layer state, consistent with category-level evidence integration in $h_t^2$ and sampling-control dynamics in $h_t^1$.}
	\label{fig23}
\end{figure}

Figure~\ref{fig23} reveals a consistent temporal progression in the upper-layer state. Across samples, early steps occupy a relatively compact region in PC space, while later steps spread out and separate by class. This pattern is consistent with progressive evidence integration under sequential glimpses. In contrast, the lower-layer state shows more heterogeneous dynamics and weaker class separation, which aligns with its role in controlling fixation sampling and maintaining local information related to gaze behavior. This analysis supports a mechanistic interpretation of EVA. The two recurrent layers capture complementary dynamics, with the upper layer emphasizing category-level integration and the lower layer emphasizing sampling control. PCA is descriptive and does not establish causality, but it provides an interpretable view of how internal states evolve during active evidence acquisition.

\section{Discussion and Conclusion}
\label{sec:conclusion}

This paper studies a recurring tension in hard-attention vision. Improving classification accuracy with stronger representations can reduce the need for informative sequential sampling, which degrades scanpath human-likeness and weakens process-level interpretability. We formalize this tension as an \emph{alignment tax} and introduce \textbf{EVA} as a mechanistic testbed that makes the performance and human-likeness trade-off measurable and adjustable under the standard classification objective. EVA uses minimal fovea--periphery sensing together with variance control and adaptive gating. The model is trained with classification labels only and does not use gaze supervision.

Across CIFAR-10, ImageNet-100, and COCO-Search18, EVA provides consistent evidence for this perspective. On CIFAR-10 with dense gaze annotations, EVA achieves a favorable accuracy--alignment trade-off among hard-attention baselines and produces scanpaths that better match human gaze under multiple metrics, including the center-debiased GCS. Mechanistic ablations isolate the roles of the components. Strong foveal features primarily drive accuracy, while variance control and gating regulate the sampling dynamics and recover human-aligned scanpaths with limited performance loss. On ImageNet-100, EVA scales to higher-resolution inputs under strict foveation. On COCO-Search18, we evaluate scanpath alignment in a gaze-supervision-free, cross-task setting. The policy is learned from classification reward on COCO and compared to human search-conditioned scanpaths without finetuning. These results suggest that mechanism-aware design can reduce the alignment tax without adding gaze supervision or changing the primary objective.

\textbf{Limitations and future work.}
First, hard-attention training remains sensitive to optimization details. More stable policy learning and variance reduction may further improve both accuracy and alignment. Second, EVA is not intended as a biologically faithful model. The modules provide functional inductive biases rather than a validated neural account. Third, scanpath evaluation is confounded by dataset biases such as center fixation. We mitigate this with a center-debiased metric, but more principled benchmarks and richer gaze data are needed. Finally, COCO-Search18 involves search-conditioned behavior, while EVA is trained on classification reward. Bridging this mismatch more directly is an important direction. Future work will improve training stability, extend EVA to broader tasks, and develop evaluation protocols that better capture human-aligned active perception.

\textbf{Acknowledgements.}
This study was funded by JST SPRING GX project (Grant Number JPMJSP2108).

\newpage
\bibliographystyle{plain}
\bibliography{references}{}

\appendix

\section{Additional EVA Inference Details}
\label{app:eva_inference}

Assigning importance to pixels is visual attention; acquiring evidence under a sensing budget is active vision. EVA is designed for the latter. Rather than processing the full image at once, EVA makes a sequence of discrete evidence requests, updating its internal state after each glimpse and selecting the next fixation conditioned on what has already been seen.

EVA instantiates this view through three interacting components: a minimal fovea--periphery observation module, a prediction-error-driven variance controller, and an adaptive gate between a lower sampling state and an upper classification state. At each glimpse step, the model receives the current fixation location, extracts a local foveal crop together with a coarse peripheral observation, encodes them into a glimpse representation, and updates the lower recurrent state. The lower state supports fixation control, while the upper state integrates gated information for classification.

The exploration scale $\sigma_t$ plays two roles during inference. First, it modulates the adaptive gate that regulates how strongly bottom-up evidence is relayed to the upper recurrent state. Second, it parameterizes the Gaussian policy used to sample the next fixation. In this way, uncertainty influences not only where the model looks next, but also how strongly newly acquired evidence is allowed to affect the evolving internal belief.

Importantly, the exploration scale is not updated inside the recurrent core itself. After the current prediction is produced, EVA computes a prediction-error signal $e_t$ and updates a state using long- and short-timescale exponential moving averages. Their discrepancy is interpreted as an uncertainty signal and mapped to the exploration scale for the next step, $\sigma_{t+1}$. Thus, the recurrent computation at step $t$ uses the current $\sigma_t$, while the prediction error observed at step $t$ determines the uncertainty level for the following decision.

Because fixation selection is stochastic, the same image can admit multiple plausible scanpaths at test time. This stochasticity is an operational form of exploration under partial observability: a single rollout records one valid trajectory of evidence acquisition, while repeated rollouts reveal the distribution of evidence requests induced by the learned policy.

Algorithm~\ref{alg:eva_inference} summarizes the inference procedure for one image. Unless otherwise stated, all results in the main paper use this standard rollout process.
\begin{algorithm}[t]
	\caption{EVA sequential inference for one image}
	\label{alg:eva_inference}
	\begin{algorithmic}[1]
		\Require image $x$, glimpse budget $T$
		
		\State initialize neuromodulator statistics
		$\bar e^{(\mathrm{long})}_0,\bar e^{(\mathrm{short})}_0$, exploration scale $\sigma_1$, initial fixation $\ell_1$, states
		$(h_0^1,c_0^1,h_0^2,c_0^2)$, 
		and gate state $\bar\beta_0$
		\For{$t=1$ to $T$}
		\State extract foveal crop $x_t^f$ and peripheral observation $x_t^p$ at fixation $\ell_t$
		\State encode fovea and periphery:
		$g_t^f \gets \phi(x_t^f)$, \quad $g_t^p \gets \psi(x_t^p,\ell_t)$
		\State form glimpse representation $s_t \gets [g_t^f \,\|\, g_t^p]$
		\State update lower recurrent state:
		$(h_t^1,c_t^1) \gets f_h^1((h_{t-1}^1,c_{t-1}^1), s_t)$
		\State compute adaptive gate using current exploration scale $\sigma_t$
		\State compute gated relay input $U_t$ from bottom-up and top-down signals
		\State update upper recurrent state:
		$(h_t^2,c_t^2) \gets f_h^2((h_{t-1}^2,c_{t-1}^2), U_t)$
		\State predict logits/probabilities $y_t \gets f_a(h_t^2)$
		\State estimate REINFORCE baseline $b_t$
		\State predict fixation mean $\mu_t \gets f_\ell(h_t^1)$
		\If{$t<T$}
		\State sample next fixation:
		$\ell_{t+1} \sim \mathcal N(\mu_t,\sigma_t^2)$
		\EndIf
		\State compute prediction-error signal $e_t$ and obtain next exploration scale $\sigma_{t+1}$
		using Eqs.~(6)--(8)
		\EndFor
		\State $\hat{y} \gets \arg\max y_T$
		\State \Return $\hat{y}, \{\ell_t\}_{t=1}^T$
	\end{algorithmic}
\end{algorithm}

\section{Dynamic Scanpaths as Sequential Evidence Acquisition}
\label{app:dynamic_scanpaths}

Visual attention scores pixels to form a saliency map. Active vision does not operate through such a map. Instead, a scanpath records how an agent acquires evidence under partial observability and a stochastic policy. The two are related, but they are not equivalent. A saliency map summarizes static spatial importance, whereas a scanpath specifies a temporal trajectory of evidence requests. The former asks where informative evidence may lie; the latter reveals how evidence is accumulated over time.

This distinction is important for interpreting EVA. In EVA, a fixation is not selected from the image alone. It is conditioned on the model's evolving internal state, including what has already been observed and how newly acquired evidence changes the model's uncertainty. As a result, the same image can admit multiple plausible scanpaths. This variability should not be interpreted as noise in the ordinary sense. Rather, it reflects the exploratory character of sequential decision-making under limited sensing.

A further consequence is that a scanpath contains information that a static saliency map cannot express directly. Later fixations depend on earlier ones, so the temporal order of evidence acquisition matters. This interpretation is consistent with the perturbation analysis in Appendix~\ref{sec:stability_details}: shuffling the order of EVA's own predicted fixation locations degrades performance even when the visited locations themselves are unchanged. In other words, not only where the model looks, but also when it looks there, contributes to the final prediction.

\subsection{Sequential Evidence Accumulation}

\begin{figure*}[t]
	\centering
	\begin{subfigure}[b]{1.0\textwidth}
		\centering
		\includegraphics[width=\linewidth]{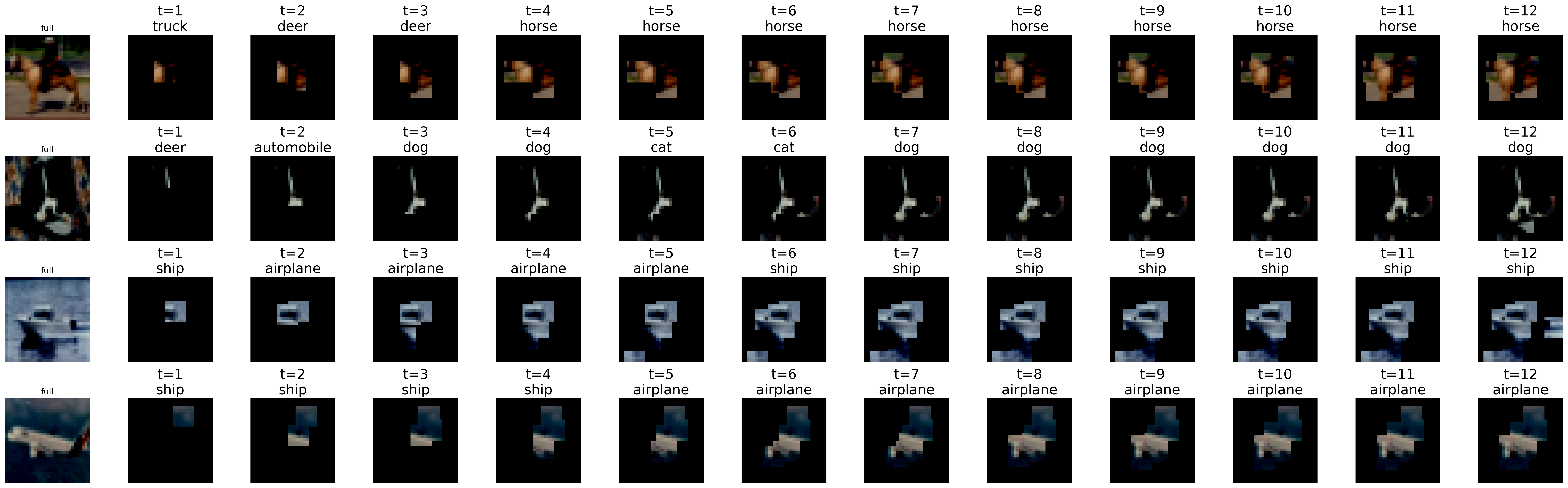}
		\caption{Foveal visible region by time}
		\label{fig84}
	\end{subfigure}\hfill
	\begin{subfigure}[b]{1.0\textwidth}
		\centering
		\includegraphics[width=\linewidth]{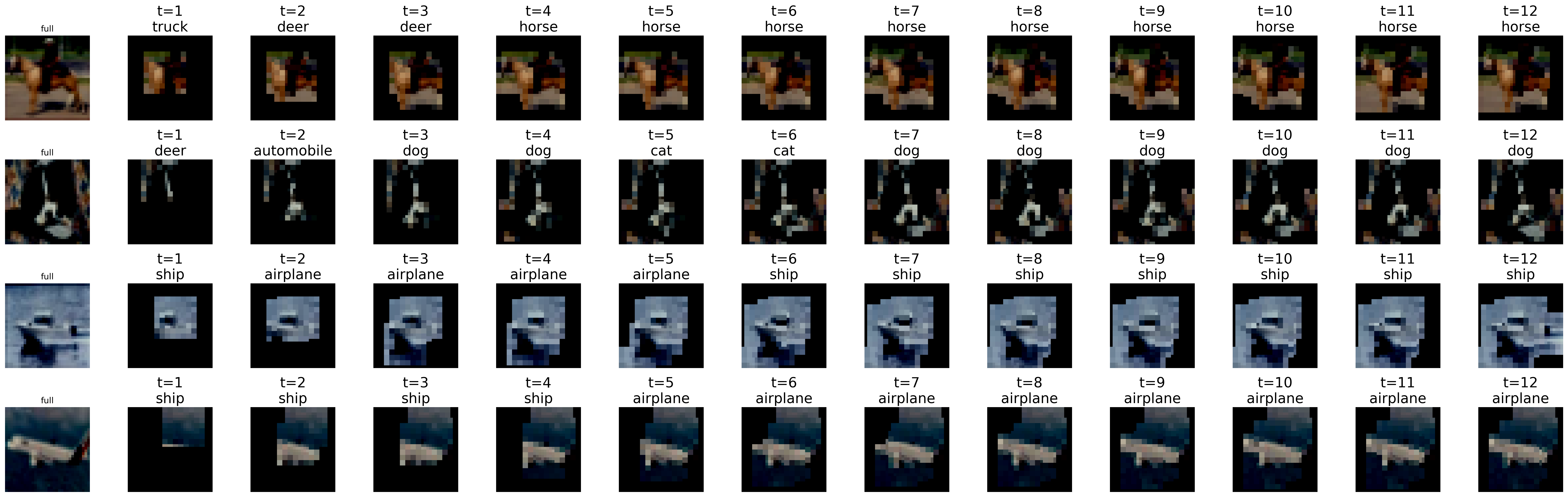}
		\caption{Peripheral visible region by time}
		\label{fig85}
	\end{subfigure}\hfill

	\caption{Sequential evidence accumulation in EVA on a representative CIFAR-10 sample. The top row shows the cumulatively observed foveal evidence over glimpse steps, and the bottom row shows the corresponding peripheral evidence. The predicted label is shown at each step, illustrating how the model's hypothesis evolves as additional evidence is acquired.}
	\label{fig:seq_accumulation}
\end{figure*}

The temporal nature of scanpaths becomes clearer when one inspects what the model has observed and predicted at each step. Figure~\ref{fig:seq_accumulation} illustrates this process on a representative example. At each glimpse step, the model has access only to the evidence gathered so far, and its prediction is conditioned on this partial observation history. As additional glimpses are acquired, the predicted label may remain stable, become more confident, or change altogether. In this sense, the scanpath is not merely a record of where the model looked, but a record of how the model's belief evolved as evidence accumulated.

In particular, a later glimpse can overturn an earlier hypothesis once previously unseen evidence becomes available. This ability to revise a prediction over time is central to active vision, but it is not represented by a static importance map alone. The sequential rollout therefore provides a process-level view of the decision, rather than only a spatial summary of potentially relevant regions.

\subsection{From Scanpaths to Fixation-Density Summaries}

\begin{figure*}[t]
	\centering
	\begin{minipage}[t]{0.24\textwidth}
		\centering
		\includegraphics[width=\linewidth]{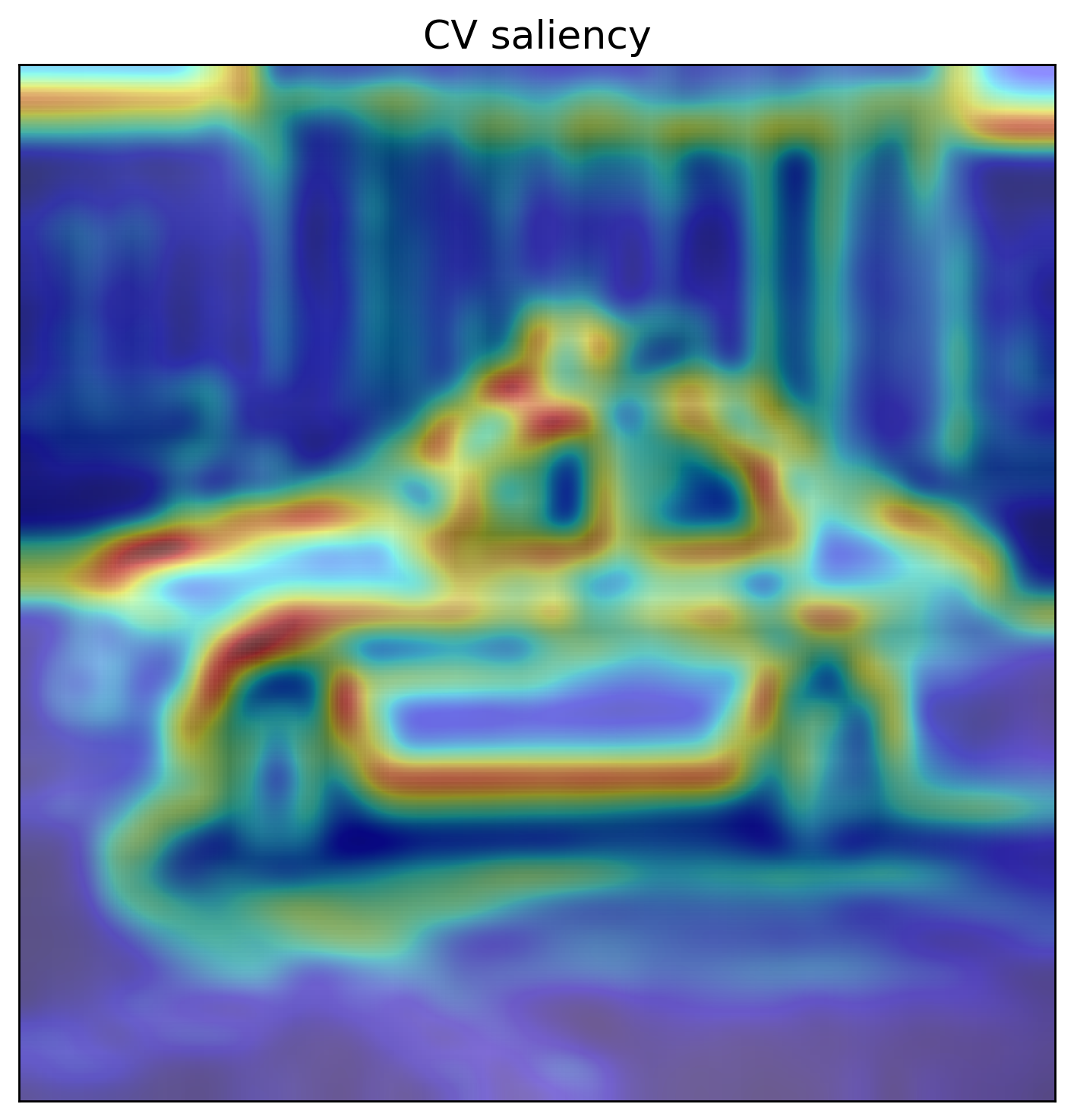}
		
		{\footnotesize (a) CV saliency}
	\end{minipage}\hfill
	\begin{minipage}[t]{0.24\textwidth}
		\centering
		\includegraphics[width=\linewidth]{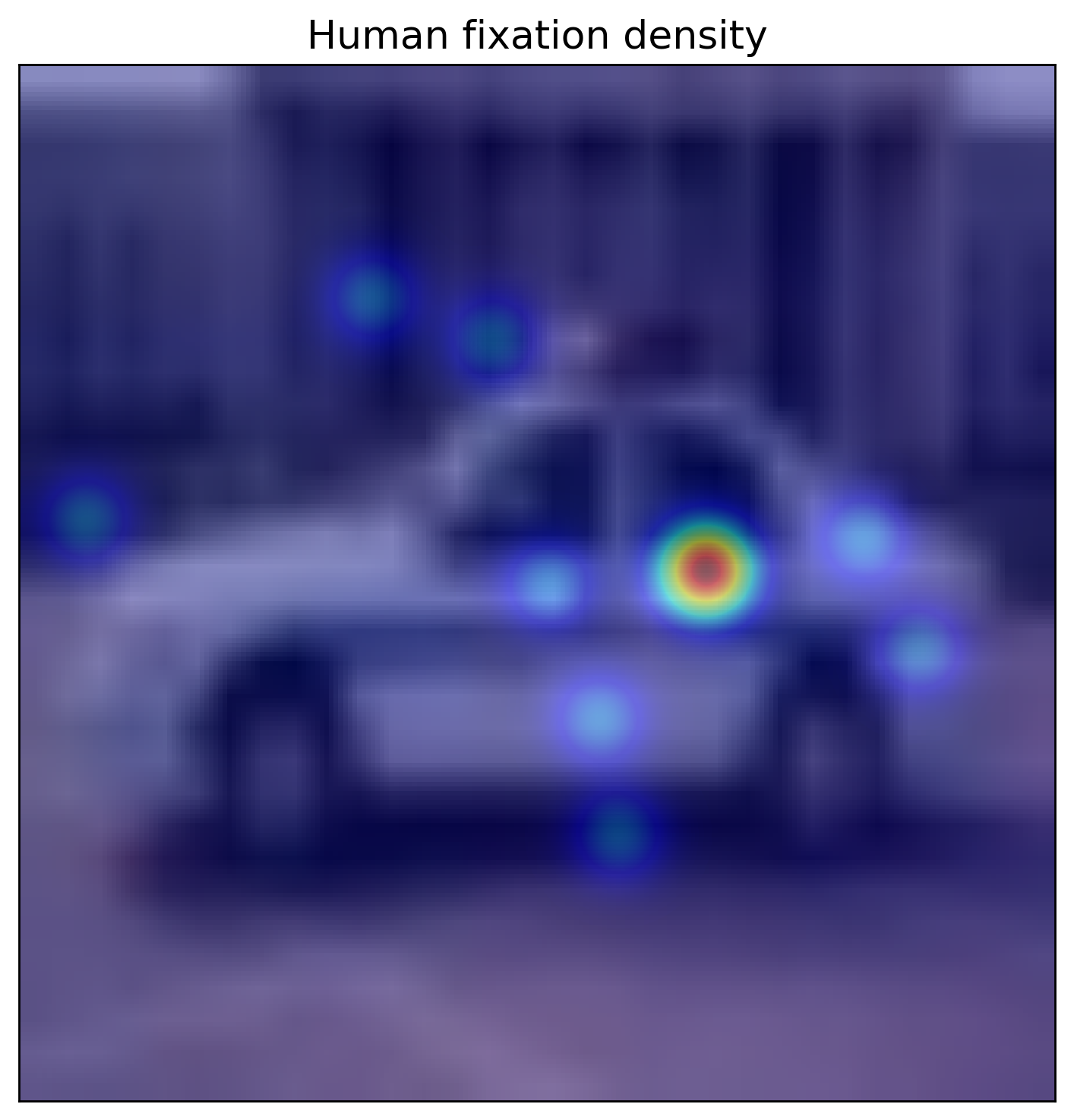}
		
		{\footnotesize (b) Human fixation density}
	\end{minipage}\hfill
	\begin{minipage}[t]{0.24\textwidth}
		\centering
		\includegraphics[width=\linewidth]{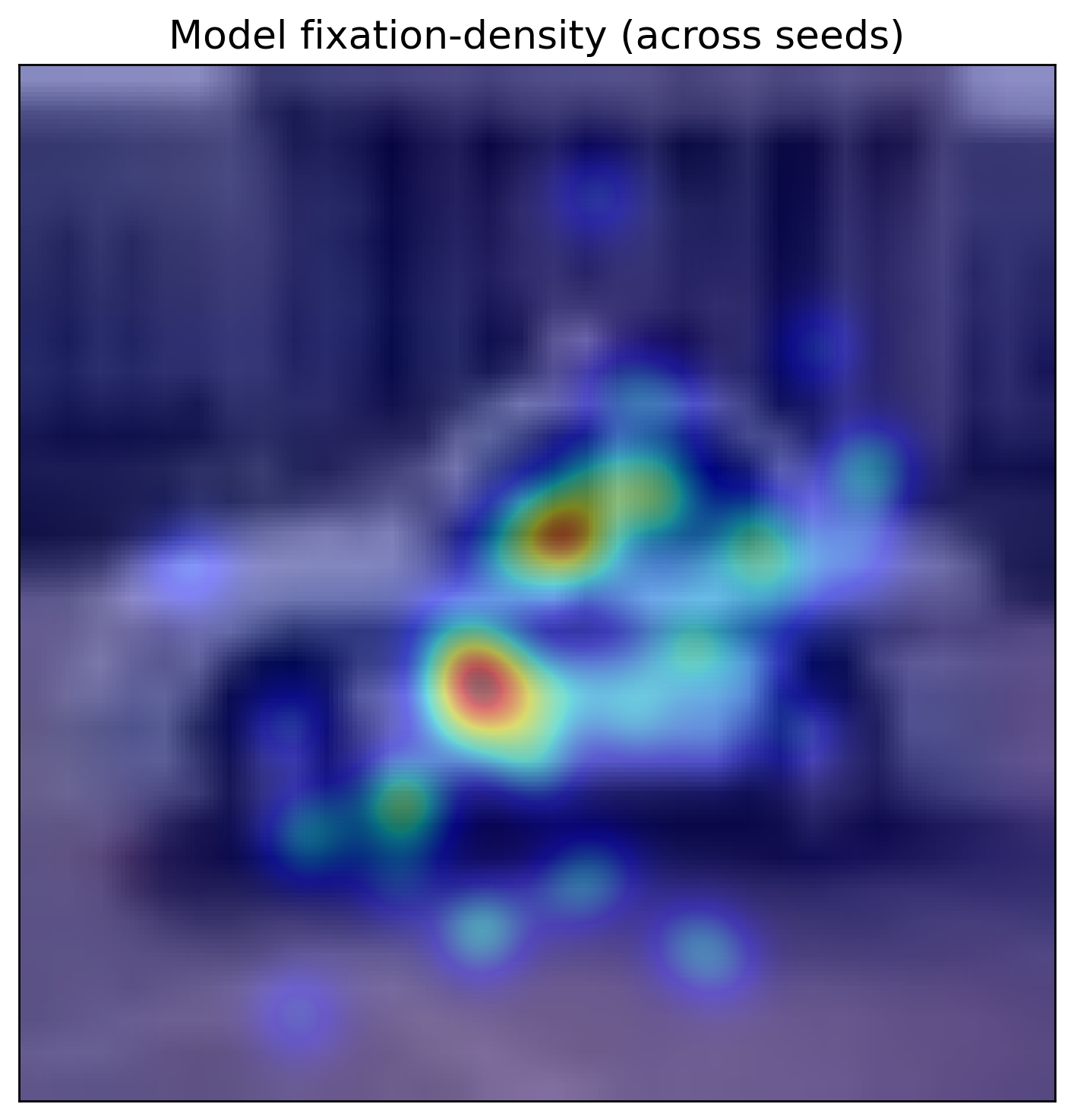}
		
		{\footnotesize (c) EVA fixation density}
	\end{minipage}\hfill
	\begin{minipage}[t]{0.24\textwidth}
		\centering
		\includegraphics[width=\linewidth]{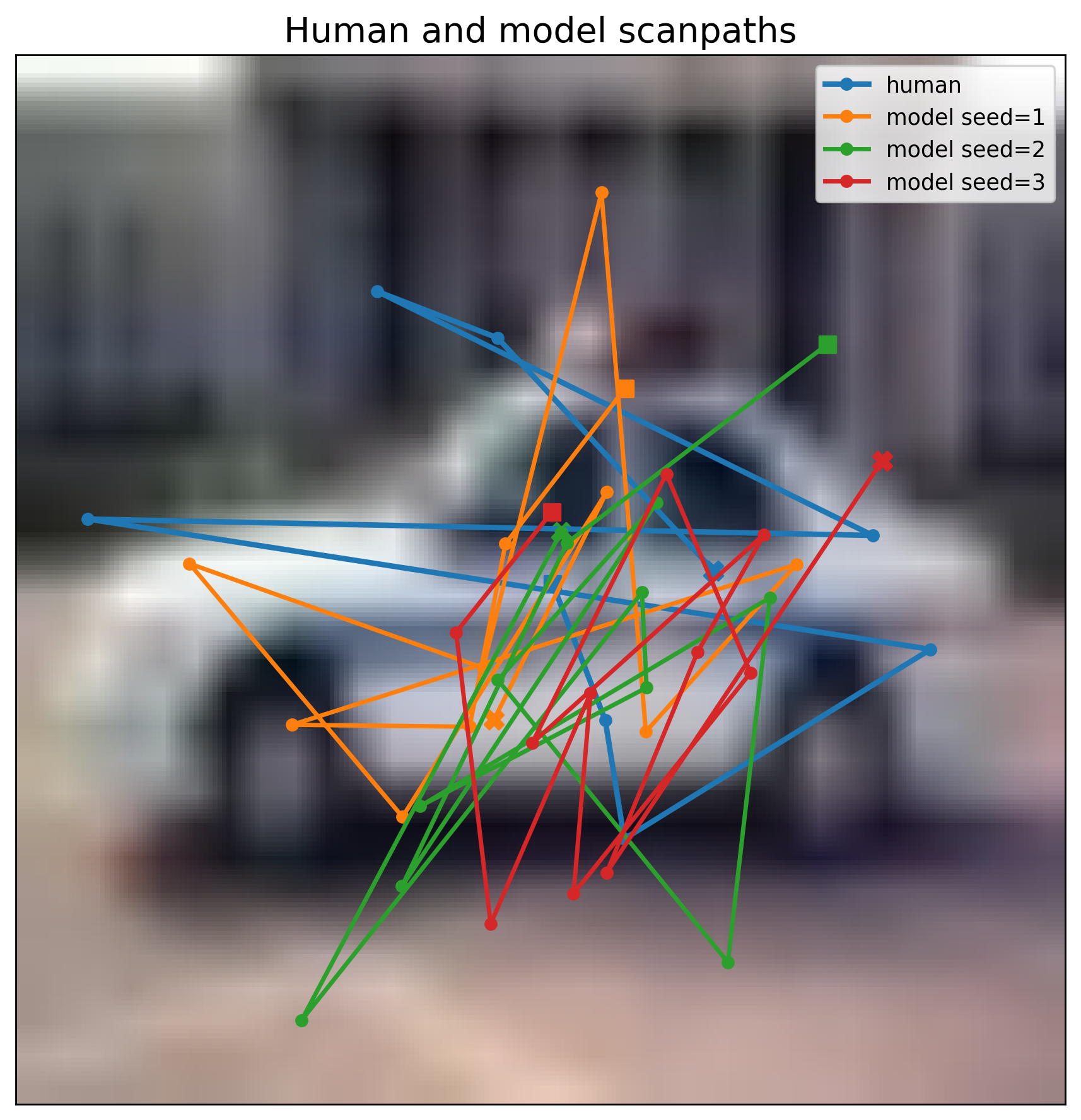}
		
		{\footnotesize (d) Human and EVA scanpaths}
	\end{minipage}
	\caption{Comparison between static saliency and scanpath-based summaries on a representative sample. (a) shows a conventional saliency map computed by a computer-vision method. (b) shows a fixation-density map reconstructed from human scanpaths. (c) shows a fixation-density map reconstructed from repeated EVA scanpaths. (d) overlays the human and EVA scanpaths on the image. The comparison highlights that static saliency and scanpath-derived summaries are related but not equivalent: saliency summarizes spatial importance, whereas scanpaths expose the sequential process of evidence acquisition.}
	\label{fig:saliency_comparison}
\end{figure*}

Although a scanpath is not itself a saliency map, repeated scanpaths can be summarized into a fixation-density representation. This provides a useful bridge between dynamic active vision and static attention visualization. Figure~\ref{fig:saliency_comparison} compares several such views on the same sample, including a conventional computer-vision saliency map, a fixation-density map reconstructed from human scanpaths, a fixation-density map reconstructed from repeated EVA scanpaths, and a direct overlay of human and model scanpaths on the image.

These views serve different purposes. A saliency map summarizes static spatial importance, whereas a scanpath and its density reconstruction reflect how evidence is sampled over time. The density map is therefore a summary of the sampling process, not a replacement for it. This distinction is important because active sensing is both sequential and stochastic: a single rollout may miss a region that appears important under an aggregated analysis, while repeated rollouts can reveal a more stable distribution of likely evidence requests.

Taken together, saliency and scanpaths should be viewed as complementary rather than competing descriptions. Saliency provides a static summary of potential importance; scanpaths expose the sequential process by which evidence is actually gathered. EVA is designed for the latter.

\section{Details of Experimental Setting}
\label{sec:model_details}

\begin{figure}[htbp]
	\centerline{\includegraphics[scale=0.75]{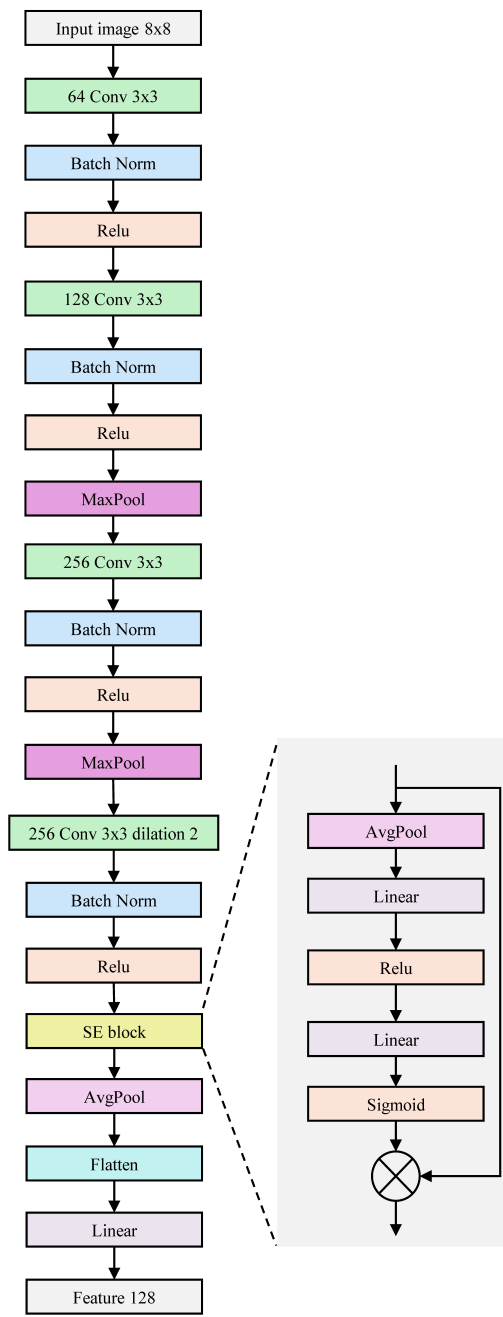}}
	\caption{CNN module used in EVA. In EVA-Large, the lightweight CNN width is increased from 64-128-256-256 to 96-192-384-384, while the remaining architectural components are kept unchanged.}
	\label{fig9}
\end{figure}

In the main classification experiments, we train the CNN baselines from scratch, without ImageNet pretraining, to enable a fair comparison with hard-attention models that do not rely on large-scale external supervision. Some hard-attention variants, however, require pretrained visual backbones by design. In particular, Saccader and EVA-Mobile use pretrained CNN modules, where the backbone is trained under the same baseline setup used in our comparisons. For Saccader, we use a ResNet backbone instead of BAGNet for a fairer comparison with the other models. In EVA, we use the lightweight CNN module shown in Fig.~\ref{fig9}. The DRAM 1-scale and 2-scale variants use the same CNN architecture as EVA. Because Saccader requires a pretrained CNN backbone, we do not evaluate it with the lightweight CNN used in EVA. For visualization, Saccader selects patches on an $8\times8$ discrete grid and orders them by score to form a pseudo-scanpath.

The ViT baseline uses a patch size of 4, hidden dimension 512, 4 transformer layers, 6 attention heads, MLP dimension 256, and dropout 0.1. For all hard-attention models, including RAM, DRAM, MRAM, and EVA, we keep the main rollout hyperparameters fixed across experiments: 12 glimpse steps and two patches per step, consisting of an $8\times8$ foveal crop and a $16\times16$ peripheral crop. When we indicate \textit{1-scale}, only the foveal crop is used and no peripheral observation is provided.

FLOPs are computed under the same testing condition using a batch of 9 images. For the hard-attention models in Table~\ref{tab:tab1}, the reported FLOPs are accumulated over the full 12-step glimpse sequence. The random seed is fixed to 1 in all experiments for reproducibility and fair comparison.

\subsection{Details of Scanpath Similarity Metrics and GCS}
\label{sec:metrics details}

We report the standard scanpath metrics DTW, ScanMatch, NSS, and AUC. 
DTW measures trajectory dissimilarity, where smaller values indicate better agreement. 
ScanMatch, NSS, and AUC measure different forms of scanpath or fixation agreement, where larger values indicate better agreement. 
Because these metrics can be inflated by center bias, our main summary metric is the center-debiased Gaze Consistency Score (GCS), following~\cite{DebiasCentralFixation}.

GCS is constructed using three calibration references for each metric: 
a human upper reference, a corner-fixation lower reference, and an always-center reference. 
The human upper reference represents the ideal agreement target. 
The corner-fixation reference serves as a practical low-reference policy by forcing all fixations to an off-target image corner, thereby removing both meaningful spatial targeting and temporal structure. 
The always-center reference is used to make center bias explicit: a model that matches human scanpaths only through central overlap should not receive an artificially high score.

For DTW, where lower is better, we first convert the raw score into a higher-is-better normalized similarity:
\begin{equation}
	\tilde{D} = \frac{D_{\min}-D}{D_{\min}-D_{\max}}.\tag{20}
\end{equation}
For the remaining metrics $M \in \{SM, NSS, AUC\}$, where larger is better, we use
\begin{equation}
	\tilde{M} = \frac{M-M_{\min}}{M_{\max}-M_{\min}}.\tag{21}
\end{equation}
We then subtract the normalized always-center reference to obtain a center-debiased score:
\begin{equation}
	\tilde{M}_{db} = \tilde{M} - \tilde{M}_{center}.\tag{22}
\end{equation}

To further discourage policies that match only spatial center overlap while failing to reproduce the movement characteristics of human scanpaths, GCS includes a movement-consistency term. 
We compute a relative-error distance
\begin{equation}
	d=\sqrt{\frac{1}{K}\sum_{k=1}^{K}
		\left(
		\frac{|f_k^{\mathrm{model}}-f_k^{\mathrm{human}}|}
		{|f_k^{\mathrm{human}}|+\epsilon}
		\right)^2},\tag{23}
\end{equation}
where $f_k$ denotes a run-level movement statistic. 
In our implementation, these statistics include total path length, mean saccade amplitude, mean distance to center, spatial coverage, direction entropy, and collapse rate. 
This distance is mapped to a movement similarity score
\begin{equation}
	\mathrm{Sim}_{move}=\exp(-d/\tau).\tag{24}
\end{equation}

The final GCS is defined as
\begin{equation}
	\mathrm{GCS}
	=
	\frac{1}{4}
	\sum_{M\in\{D,SM,NSS,AUC\}}
	\tilde{M}_{db}
	+
	\lambda\,\mathrm{Sim}_{move},\tag{25}
\end{equation}
where $\lambda=0.1$ in all experiments. 
Thus, GCS rewards alignment with human scanpaths only when the model exceeds the center baseline and also exhibits compatible movement dynamics.

In practice, this debiasing is important because several learned policies can appear competitive under DTW, ScanMatch, NSS, or AUC while remaining close to the always-center reference. 
GCS makes this confound explicit by calibrating each metric against the human, corner-fixed, and center-fixed references before aggregation.
All scanpath metrics are computed after normalizing model and human coordinates to the same reference frame, and GCS is evaluated on the same image subset used for human-model scanpath comparison.

\subsection{Details of Stability Test}
\label{sec:stability_details}

To assess the functional role of EVA's learned gaze policy, we evaluate the model under several controlled perturbations of the fixation sequence.

\textbf{Center-fixed} forces every glimpse to the image center at every step. This removes learned spatial exploration and tests how much performance can be supported by central bias alone.

\textbf{Corner-fixed} deterministically fixes every glimpse to one image corner (top-right in our experiments), regardless of the input. This probes robustness under systematic spatial misalignment with the object of interest.

\textbf{Random} samples each glimpse independently from a uniform distribution over the image plane at every step. This preserves the glimpse budget while discarding the learned spatial policy.

\textbf{Shuffled} uses EVA's own predicted fixation locations but randomly permutes their temporal order. This preserves the exact set of attended points while destroying the order in which they are visited.

Together, these perturbations disentangle different contributions of the learned policy: central bias, spatial targeting, and temporal ordering. Table~\ref{tab:2} summarizes the corresponding results. EVA degrades substantially under corner-fixed and shuffled policies, indicating that its performance depends not only on where it looks, but also on the temporal structure of the scanpath. By contrast, RAM and DRAM degrade strongly under random or corner-fixed policies but are nearly unaffected by shuffling, suggesting that their learned policies are less tightly coupled to fixation order. This difference supports the view that EVA's scanpaths are functionally meaningful rather than merely decorative visualizations.

\begin{table}[t]
	\centering
	\caption{Accuracy (\%) of hard-attention models under different gaze policies on CIFAR-10. The \emph{Predicted} column reports the learned policy; the remaining columns replace or perturb the gaze trajectory. Numbers in parentheses indicate the accuracy drop relative to \emph{Predicted}.}
	\label{tab:2}
	\begin{tabular}{lccccc}
		\toprule
		\multirow{2}{*}{Model} & \multicolumn{5}{c}{Gaze Policy} \\
		\cmidrule(lr){2-6}
		& Predicted & Center-fixed & Corner-fixed $\downarrow$ & Random & Shuffled $\downarrow$ \\
		\midrule
		RAM           & 61.6 & 45.1 (-16.5) & 12.8 (-48.8) & 43.5 (-18.1) & 61.3 (-0.3) \\
		DRAM          & 62.2 & 41.6 (-20.6) & 11.6 (-50.6) & 38.4 (-23.8) & 61.9 (-0.3) \\
		MRAM          & 58.3 & 44.1 (-14.2) & 13.6 (-44.7) & 48.7 (-9.6)  & 56.9 (-1.4) \\
		\textbf{EVA}  & 79.8 & 57.6 (-22.2) & 19.6 (-60.2) & 70.5 (-9.3)  & 71.3 (-8.5) \\
		\bottomrule
	\end{tabular}
\end{table}

\paragraph{Stability under gaze perturbations.}
The perturbation results provide a process-level probe of interpretability. Constraining EVA to center-only, corner-only, random, or shuffled trajectories reduces performance in distinct ways, revealing which aspects of the learned policy are functionally important. In particular, EVA's sensitivity to shuffled trajectories shows that the order of evidence acquisition matters even when the visited locations themselves are unchanged. This supports the interpretation that EVA's scanpaths encode a meaningful sequential strategy for recognition.

\section{Additional Diagnostic Analyses}
\label{sec:diagnostic_analyses}

The analyses in this section probe EVA at test time without changing the trained model parameters. Rather than introducing new training objectives, they modify only the conditions under which the learned policy is executed. This allows us to examine whether EVA's behavior is robust, interpretable, and diagnostically informative at the level of sequential evidence acquisition.

\subsection{Sensitivity to Test-Time Initialization}
\label{sec:init_sensitivity}

Because EVA is a sequential and stochastic policy, its early behavior can depend on how the rollout is initialized. We therefore examine the sensitivity of a fixed trained checkpoint to different test-time initial conditions, varying the initial fixation policy and the initial recurrent state while keeping all learned parameters unchanged.

\begin{figure*}[t]
	\centering
	\begin{minipage}[t]{0.48\textwidth}
		\centering
		\includegraphics[width=\linewidth]{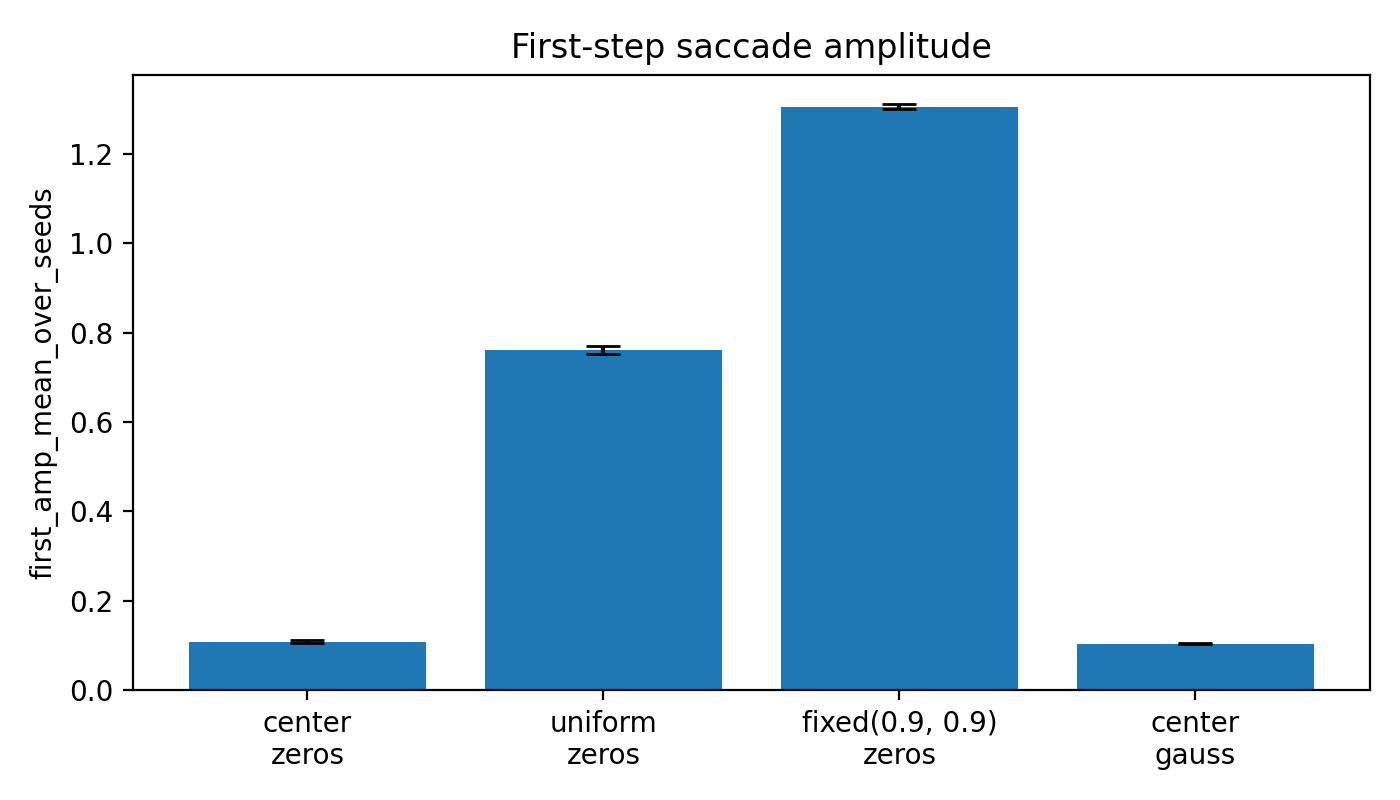}
	\end{minipage}\hfill
	\begin{minipage}[t]{0.48\textwidth}
		\centering
		\includegraphics[width=\linewidth]{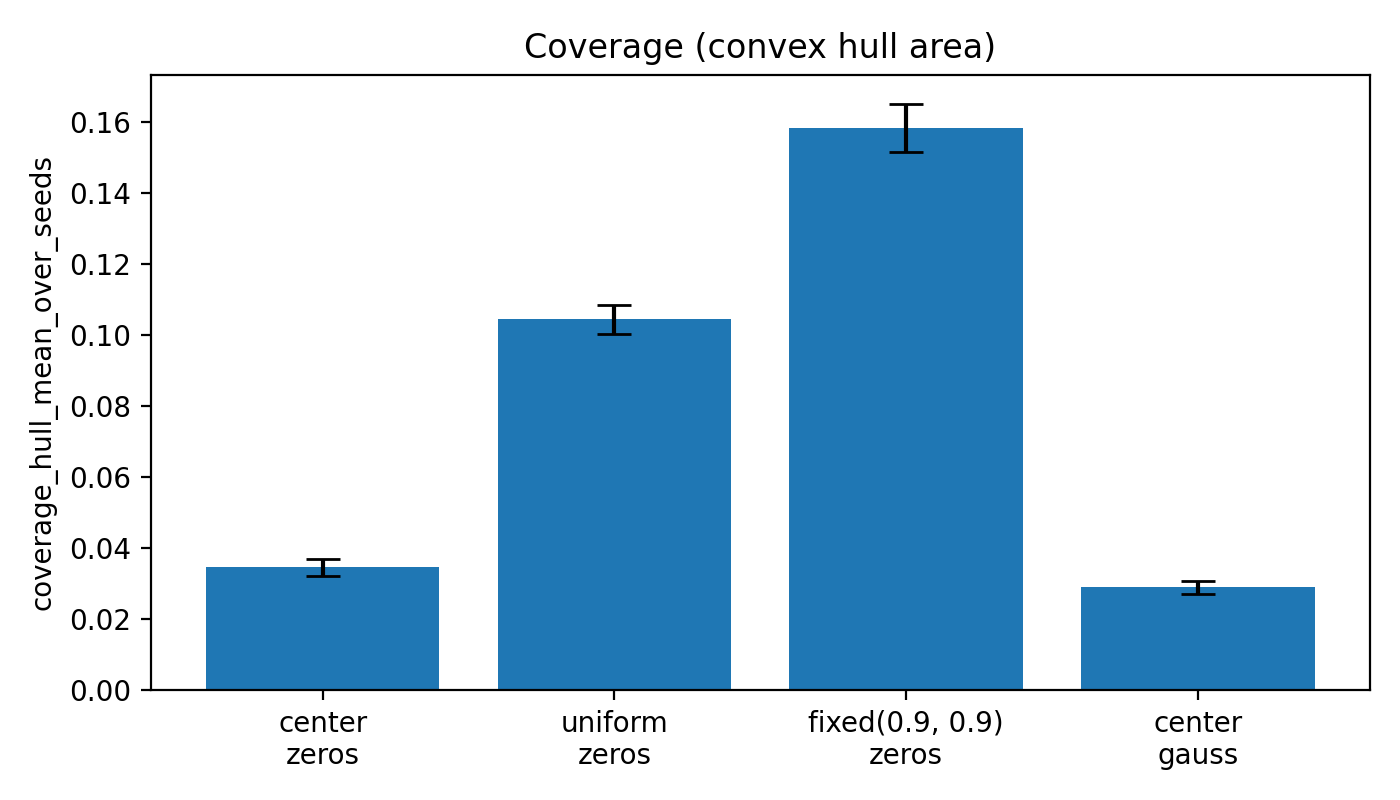}
	\end{minipage}
	
	\vspace{0.5em}
	
	\begin{minipage}[t]{0.48\textwidth}
		\centering
		\includegraphics[width=\linewidth]{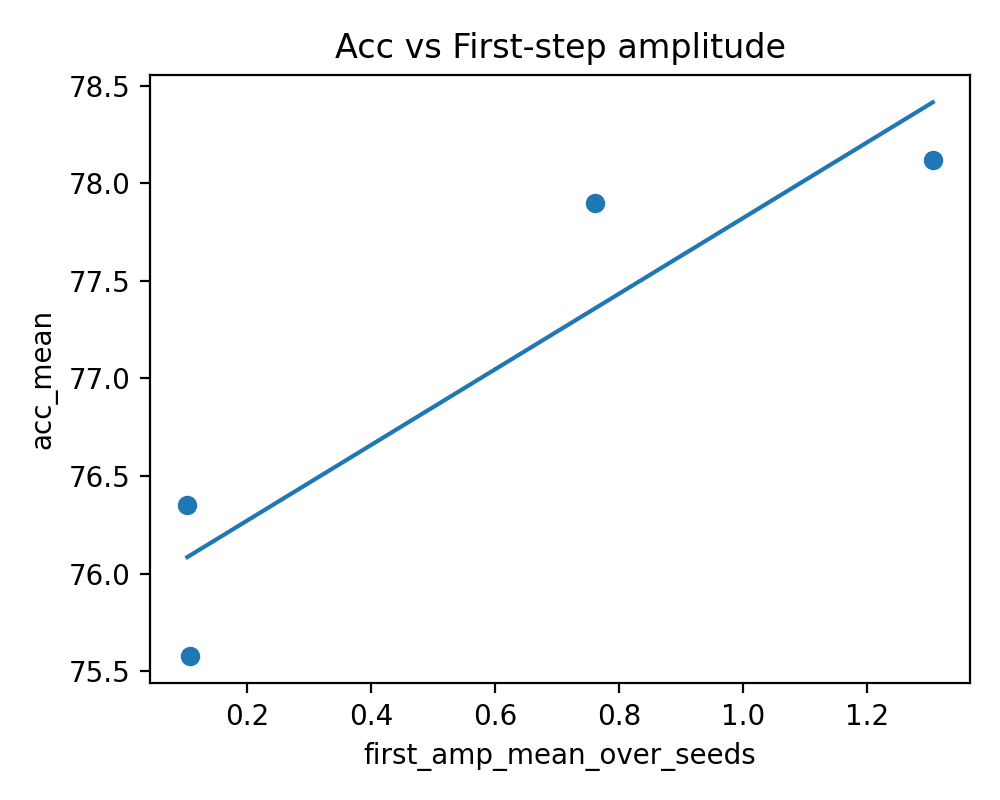}
	\end{minipage}\hfill
	\begin{minipage}[t]{0.48\textwidth}
		\centering
		\includegraphics[width=\linewidth]{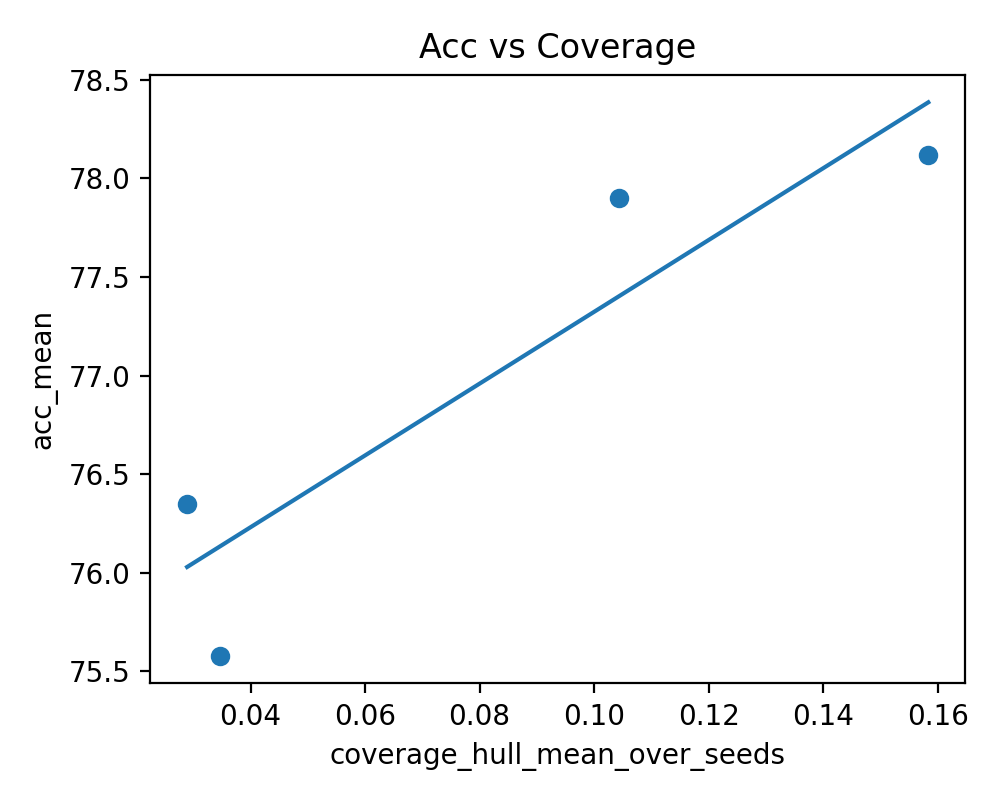}
	\end{minipage}
	\caption{Sensitivity of a fixed EVA checkpoint to test-time initialization. Top: different initialization conditions induce different levels of early exploration, as measured by first-step saccade amplitude and spatial coverage. Bottom: across the tested initialization conditions, broader early exploration is associated with higher classification accuracy. These plots are intended as diagnostic summaries rather than as a statistical claim about population-level correlation.}
	\label{fig:init_sensitivity}
\end{figure*}

Figure~\ref{fig:init_sensitivity} summarizes the resulting behavior. Different initial conditions induce clear and interpretable changes in early exploration. Off-center initialization leads to larger first-step saccades and broader spatial coverage, whereas center-based initialization yields more localized trajectories. These differences are accompanied by modest but systematic changes in classification accuracy.

Importantly, the effect is structured rather than erratic. The same initialization changes that increase early exploration also tend to improve task performance, suggesting that EVA's test-time sensitivity is not merely noise injection but a controlled change in how evidence acquisition begins. This supports the view that EVA behaves as an active sensing policy whose predictions depend on both the current image and the scanpath by which evidence is gathered.

\subsection{Human-Guided Intervention as a Diagnostic Direction}
\label{sec:human_guided_direction}

A natural extension of the above analysis is to intervene directly on the fixation sequence at test time, for example by replacing part of EVA's predicted scanpath with human fixation locations while keeping all trained parameters fixed. Such an intervention would not introduce gaze supervision into training; rather, it would provide a direct diagnostic probe of whether part of the residual performance gap is attributable to evidence acquisition rather than to the classifier alone. We regard this as a promising direction for future analysis, but defer a systematic evaluation to future work.

\section{Alignment of CIFAR-10 and Gaze-CIFAR-10}
\label{sec:gaze_alignment}

This section describes how we align the original CIFAR-10 images with the corresponding human gaze recordings from Gaze-CIFAR-10, and how we convert raw gaze traces into sequential fixation sequences comparable to model-predicted scanpaths.

\subsection{Image Alignment}

The human gaze data were recorded on upsampled versions of the original CIFAR-10 images at a resolution of $1024\times1024$. Because the Gaze-CIFAR-10 data did not provide an explicit one-to-one mapping to the original CIFAR-10 test images, we established correspondence using perceptual image hashing.

Specifically, we constructed a reference database from all 10,000 images in the CIFAR-10 test split. For each original image, we computed a perceptual hash (pHash) and stored the corresponding image index and class label. For each gaze image, we first downsampled the $1024\times1024$ image back to the original $32\times32$ resolution using bilinear interpolation, then computed its pHash and matched it against the reference database. Exact matches were used whenever available; otherwise, we selected the nearest reference hash under Hamming distance to account for small differences introduced by upsampling and downsampling.

To verify that the recovered matches were reliable, we additionally performed a manual spot check by randomly sampling aligned pairs across all CIFAR-10 classes and visually confirming the correspondence between the original and upsampled images.

\subsection{Sequential Fixation Extraction}

The raw gaze recordings contain dense gaze-point sequences, often with hundreds of coordinate samples per image. To obtain human scanpaths that are comparable to the fixed-length scanpaths produced by EVA, we converted these raw traces into sequential fixation sequences.

After alignment, gaze coordinates were normalized to the common reference frame used in our scanpath evaluation. We then applied an I-DT (Identification by Dispersion Threshold) style procedure to cluster gaze points into fixations. Gaze samples falling within a spatial radius of 15 pixels were grouped into a fixation cluster, and the center of each cluster was taken as the fixation location. Finally, we retained exactly 12 sequential fixations per gaze recording to match the 12-step rollout used by the hard-attention models.

This procedure yields human scanpaths that are aligned with the original CIFAR-10 images in both image identity and fixation sequence length. As a result, human and model scanpaths can be compared consistently in space and time using the scanpath similarity metrics and GCS defined in Appendix~\ref{sec:metrics details}.

\subsection{Use in Evaluation}

The aligned Gaze-CIFAR-10 subset is used only for human--model scanpath comparison. Model classification results are still reported on the standard CIFAR-10 benchmark, while human gaze is recorded on $1024\times1024$ images and model scanpaths are generated from the corresponding $32\times32$ CIFAR-10 images. For scanpath evaluation, both human and model coordinates are mapped to a common $224\times224$ reference frame. This separation ensures that the scanpath analysis reflects alignment with human evidence-acquisition behavior rather than a change in the underlying classification protocol.

\section{Additional Qualitative Examples}
\label{sec:qual_examples}

This section provides additional qualitative examples of EVA's sequential decision process. Rather than comparing exact trajectory overlap with human scanpaths, we focus on how EVA's predictions evolve as evidence is accumulated over time. The examples illustrate two complementary regimes: successful evidence revision, in which later glimpses correct an earlier hypothesis, and failure cases, in which the policy acquires only partially diagnostic evidence and converges to an incorrect class.

\subsection{Successful Evidence Revision}

\begin{figure*}[t]
	\centering
	\begin{minipage}[t]{0.98\textwidth}
		\centering
		\includegraphics[width=\linewidth]{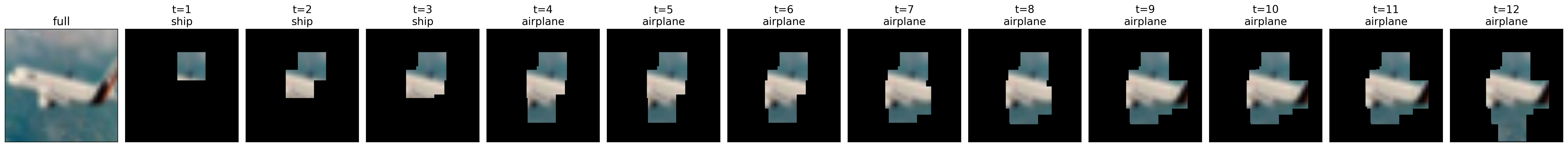}
		\vspace{2pt}
		
		{\footnotesize (a) Success case 1: \textit{ship} $\rightarrow$ \textit{airplane}}
	\end{minipage}\hfill
	\begin{minipage}[t]{0.98\textwidth}
		\centering
		\includegraphics[width=\linewidth]{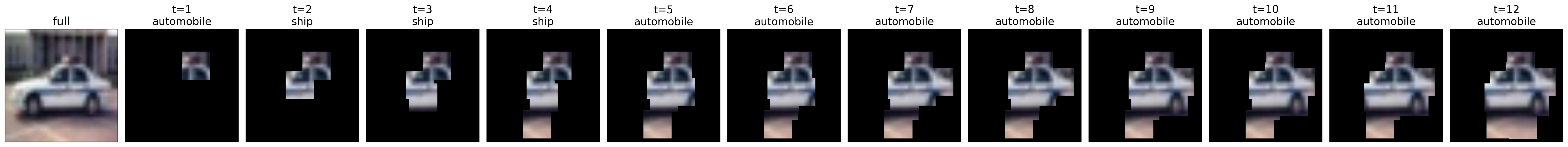}
		\vspace{2pt}
		
		{\footnotesize (b) Success case 2: \textit{automobile} $\rightarrow$ \textit{ship} $\rightarrow$ \textit{automobile}}
	\end{minipage}
	\caption{Representative examples of successful evidence revision in EVA. In (a), the model initially favors \textit{ship} but switches to \textit{airplane} once later glimpses reveal more diagnostic object structure. In (b), the model briefly drifts toward \textit{ship} under partial evidence, then returns to the correct \textit{automobile} hypothesis after more of the vehicle body becomes visible. These examples show that EVA's prediction is shaped by the temporal order in which evidence is acquired.}
	\label{fig:qual_success}
\end{figure*}

Figure~\ref{fig:qual_success} shows two representative successful cases in which EVA revises its prediction as additional evidence is gathered. In the first example, the model initially predicts \textit{ship}, likely because the earliest glimpses contain mostly background and only a weak local cue from the object. Once the rollout expands to include the elongated fuselage and wing structure, the prediction switches to \textit{airplane} and remains stable thereafter. In the second example, the model begins with the correct coarse hypothesis \textit{automobile}, briefly drifts toward \textit{ship} when only a limited local region is visible, and then returns to \textit{automobile} after later glimpses reveal more of the vehicle body and its overall shape. These examples illustrate that EVA does not merely accumulate evidence monotonically; its internal belief can be revised as the visible structure changes over time.

\subsection{Failure Cases}

\begin{figure*}[t]
	\centering
	\begin{minipage}[t]{0.98\textwidth}
		\centering
		\includegraphics[width=\linewidth]{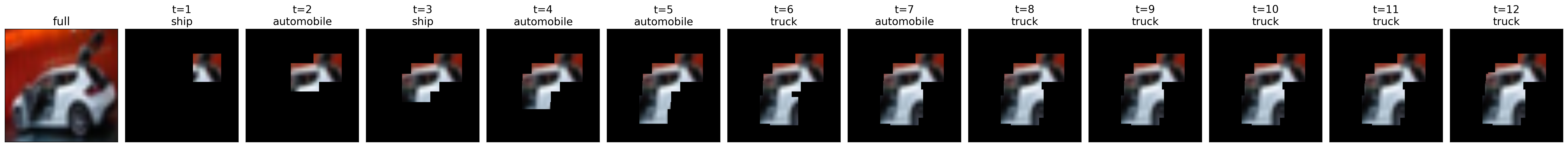}
		\vspace{2pt}
		
		{\footnotesize (a) Failure case 1: unstable vehicle hypothesis, final \textit{truck}}
	\end{minipage}\hfill
	\begin{minipage}[t]{0.98\textwidth}
		\centering
		\includegraphics[width=\linewidth]{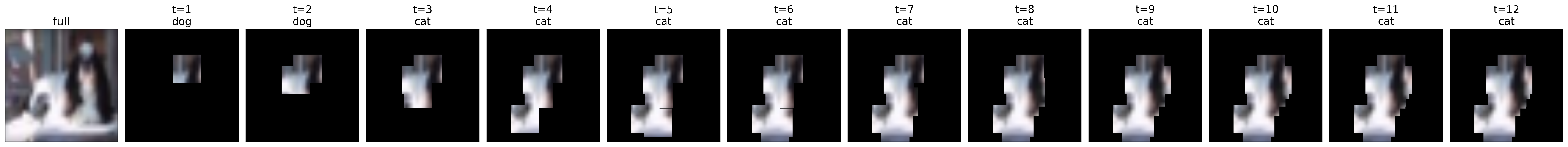}
		\vspace{2pt}
		
		{\footnotesize (b) Failure case 2: \textit{dog} $\rightarrow$ \textit{cat}}
	\end{minipage}
	\caption{Representative failure cases of EVA. In (a), the model eventually commits to \textit{truck} after observing evidence sufficient for a coarse vehicle category but insufficient for finer discrimination. In (b), the model revises from \textit{dog} to \textit{cat} and then remains trapped in that hypothesis, suggesting that later glimpses reinforce a locally plausible but ultimately misleading interpretation. These failures remain interpretable at the level of the scanpath and the accumulated evidence.}
	\label{fig:qual_failure}
\end{figure*}

Figure~\ref{fig:qual_failure} shows two representative failure cases. In the first example, the model gradually settles on \textit{truck} after an initially unstable sequence of \textit{ship}/\textit{automobile} hypotheses. The rollout captures enough evidence to identify the object as a road vehicle, but not enough fine-grained structure to resolve the distinction in favor of the the correct automobile class. In the second example, the model shifts from \textit{dog} to \textit{cat} early in the rollout and remains committed to that hypothesis. Here, the glimpses repeatedly sample a locally plausible but only partially diagnostic region, while failing to acquire stronger evidence that would distinguish the correct class. In both cases, the final error is interpretable as a failure of sequential evidence acquisition rather than an opaque classifier output.

\subsection{Different Rollout Seeds, Different Evidence Paths}

\begin{figure*}[t]
	\centering
	\begin{minipage}[t]{0.98\textwidth}
		\centering
		\includegraphics[width=\linewidth]{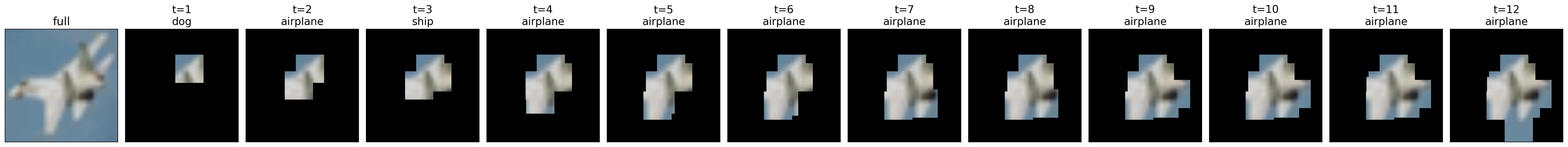}
		\vspace{2pt}
		
		{\footnotesize (a) Rollout seed 1: converges to \textit{airplane}}
	\end{minipage}\hfill
	\begin{minipage}[t]{0.98\textwidth}
		\centering
		\includegraphics[width=\linewidth]{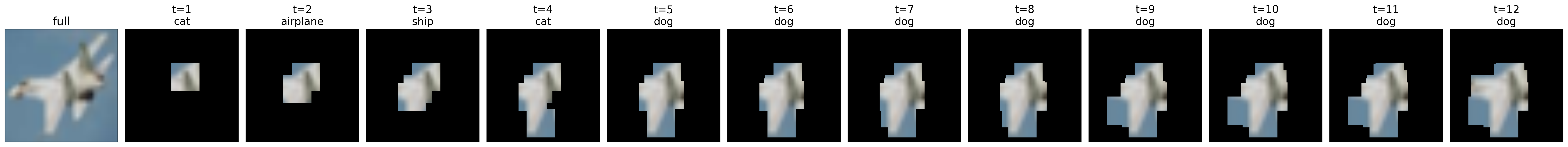}
		\vspace{2pt}
		
		{\footnotesize (b) Rollout seed 2: converges to \textit{dog}}
	\end{minipage}
	\caption{Same image under two different stochastic rollouts. In (a), EVA initially fluctuates among several hypotheses but eventually converges to the correct \textit{airplane} class once later glimpses reveal more diagnostic structure. In (b), a different rollout remains unstable under ambiguous local evidence and ultimately converges to the incorrect \textit{dog} class. This example illustrates that different scanpaths can induce different evidence-acquisition scanpaths and therefore different prediction outcomes.}
	\label{fig:qual_seed_pair}
\end{figure*}

Because EVA uses a stochastic fixation policy, the same image can admit multiple plausible scanpaths at test time. These different trajectories do not merely change the visual appearance of the rollout; they can also lead to different prediction outcomes. Figure~\ref{fig:qual_seed_pair} shows a representative example in which two rollout seeds produce different evidence-acquisition scanpaths for the same image.

In the top example, the model initially fluctuates among several hypotheses but eventually converges to the correct \textit{airplane} class once later glimpses reveal sufficiently diagnostic object structure. In the bottom example, a different rollout remains unstable under ambiguous local evidence and ultimately converges to the incorrect \textit{dog} class. The comparison highlights that EVA's prediction depends not only on the image itself, but also on the scanpaths by which evidence is sampled and integrated over time.

This example should not be interpreted as arbitrary randomness. Rather, it illustrates the scanpaths dependence of sequential evidence acquisition: different stochastic trajectories can expose different subsets of evidence, and these differences can push the recurrent state toward either a correct or an incorrect interpretation.

\paragraph{Interpretation.}
Taken together, these examples reinforce the main qualitative picture of EVA. When EVA succeeds, it succeeds by revising its internal belief as new evidence becomes available. When it fails, the failure can often be traced to incomplete or misleading sequential evidence, such as missing a discriminative region or overcommitting to a partially supported hypothesis. The paired rollout example further shows that, even for the same image, different stochastic evidence-acquisition scanpaths can lead to different outcomes. This process-level interpretability is precisely what distinguishes scanpath-based active vision from static importance summaries alone.


\newpage

\end{document}